\documentclass{article}

\PassOptionsToPackage{numbers, compress}{natbib}
\usepackage[preprint]{neurips_2026}

\usepackage[utf8]{inputenc} 
\usepackage[T1]{fontenc}    
\usepackage{hyperref}       
\usepackage{url}            
\usepackage{booktabs}       
\usepackage{amsfonts}       
\usepackage{nicefrac}       
\usepackage{microtype}      
\usepackage{xcolor}         
\usepackage{amsmath}
\usepackage{amssymb}
\usepackage{mathtools}
\usepackage{amsthm}

\usepackage{graphicx}
\usepackage{amsmath}
\usepackage{booktabs}       
\usepackage{amsfonts}       
\usepackage{nicefrac}       
\usepackage{microtype}      
\usepackage{xcolor}         
\usepackage{wrapfig}
\usepackage{varwidth}
\usepackage{subcaption}
\usepackage{adjustbox}
\usepackage{colortbl}
\usepackage{multirow}
\usepackage{algorithm}
\usepackage{algorithmicx}
\usepackage{algpseudocode}
\DeclareMathOperator*{\UniformSample}{UniformSample}
\DeclareMathOperator*{\UniformInt}{UniformInt}
\DeclareMathOperator*{\TopKIndices}{TopKIndices}
\title{Elastic MoE: Unlocking the Inference-Time Scalability of Mixture-of-Experts}
\vspace{-0.5cm}

\author{
Naibin Gu\textsuperscript{\rm 1,2}\thanks{~ denotes equal contribution. $^\dagger$ denotes corresponding author.}\ ,\ Zhenyu Zhang\textsuperscript{\rm 3}\footnotemark[1]\ ,\ Yuchen Feng\textsuperscript{\rm 1,2},\ Yilong Chen\textsuperscript{\rm 1,2},\ Peng Fu\textsuperscript{\rm 1,2}\ ,\ Zheng Lin\textsuperscript{\rm 1,2},\\ {\bf Shuohuan Wang\textsuperscript{\rm 3},\ Yu Sun\textsuperscript{\rm 3},\ Hua Wu\textsuperscript{\rm 3},\ Weiping Wang\textsuperscript{\rm 1},\ Haifeng Wang\textsuperscript{\rm 3}} \\ 
\textsuperscript{\rm 1}Institute of Information Engineering, Chinese Academy of Sciences, Beijing, China \\
\textsuperscript{\rm 2}School of Cyber Security, University of Chinese Academy of Sciences, Beijing, China \\
\textsuperscript{\rm 3}Baidu Inc., Beijing, China \\
  \texttt{\textrm{\{}gunaibin,fupeng\textrm{\}}@iie.ac.cn} \\
  \texttt{\textrm{\{}zhangzhenyu07,wangshuohuan\textrm{\}}@baidu.com}
}

\begin{document}

\maketitle

\vspace{-0.5cm}
\begin{abstract}
Mixture-of-Experts (MoE) models typically fix the number of activated experts $k$ at both training and inference. However, real-world deployments often face heterogeneous 
hardware, fluctuating workloads, and diverse quality-latency 
requirements, while training separate models for each 
scenario is costly. Considering that MoE models already operate with sparse 
activation, adjusting the number of activated experts offers 
a natural path to serving diverse budgets with a single model. 
Yet, we find that activating more experts $k'$ ($> k$) at 
inference does not yield the expected gains. Instead, 
performance degrades rapidly after only a slight increase, 
a phenomenon we term the \textit{inference-time scaling wall}. Further investigation reveals that this degradation stems from a lack of learned collaboration among experts. To address this, we introduce \textbf{Elastic Mixture-of-Experts (EMoE)}, a novel training framework that enables MoE models to elastically vary the number of activated experts at inference. By simultaneously training experts to collaborate in diverse combinations and encouraging the router to make high-quality selections, EMoE ensures robust performance across inference budgets.
Extensive experiments across four MoE architectures (7B--21B) 
and nine benchmarks show that EMoE significantly expands the effective scaling range to 2-3$\times$ the training-time $k$, while also achieving higher peak performance.
\end{abstract}
\vspace{-0.1cm}
\section{Introduction}
Large-scale models based on the Transformer architecture~\citep{DBLP:conf/nips/VaswaniSPUJGKP17} have demonstrated remarkable performance across a wide range of tasks~\citep{OpenAI2023GPT4TR,Touvron2023LLaMAOA,DBLP:journals/corr/abs-2307-09288}. However, this performance gain is often accompanied by a substantial increase in model size, leading to prohibitive computational costs for both training and inference. To address this challenge, the Mixture-of-Experts (MoE) paradigm~\citep{DBLP:journals/jmlr/FedusZS22,DBLP:conf/iclr/LepikhinLXCFHKS21} has garnered significant attention. By employing a sparsely activated architecture, MoE models effectively maintain model capacity while enhancing computational efficiency, leading to its widespread adoption. 

Most MoE models~\citep{DBLP:journals/corr/abs-2405-04434,qwen_moe,DBLP:conf/acl/DaiDZXGCLZYWXLH24,kimiteam2025kimik2openagentic} are implemented via a Top-$k$ strategy~\citep{DBLP:conf/iclr/ShazeerMMDLHD17}, where the number of activated experts $k$ is fixed during pretraining and kept unchanged at inference. However, in practice, the computational resources available at inference can vary significantly. For example, a model may be deployed on resource-constrained devices with tight latency budgets, or on high-end GPU clusters where ample compute is available. Similarly, fluctuating workloads and diverse quality-latency requirements in production systems make elastic inference highly desirable, yet training separate models for each scenario is costly. Given that MoE models already operate with sparse activation, making them inherently efficient under limited compute, a natural idea is to activate more experts $k'$ (where $k' > k$) when additional compute is available. This prompts a practical question: \textit{Can we unlock the latent potential of a MoE model by activating more experts when more inference compute is available?}

Unfortunately, we find that standard MoE models are unable to support such elastic inference (Section~\ref{sec:pilot}). We uncover an intriguing and previously under-explored \textbf{inference-time scaling wall}: when a model is trained with $k$ experts, the effective scaling range at inference is so narrow that increasing this to a slightly larger $k'$ ($> k$) causes performance to degrade rapidly. While increasing the number of activated experts during training offers some partial relief, it incurs prohibitive computational overhead, and we observe that such models collapse once inference returns to the original $k$ budget.
Upon further analysis, we identify the root cause of this inability to extrapolate to larger $k'$ values as disparities in \emph{expert co-occurrence frequencies}. Specifically, the additionally activated experts at inference have not been trained to collaborate effectively with the originally selected experts, as these new combinations are rarely encountered during training. This lack of learned collaboration causes the observed performance degradation.

In this paper, we introduce \textbf{Elastic Mixture-of-Experts (EMoE)}, a novel lightweight training framework that equips MoE models with inference-time elasticity. Without any architectural modification, EMoE can be applied in a plug-and-play manner during post-training on pretrained MoE checkpoints, enabling a single model to scale up to more experts when compute permits and maintain strong performance with fewer experts under low inference budgets.
The effectiveness of EMoE stems from two key designs. First, to address collaboration failure, we propose \textbf{stochastic co-activation sampling}, which draws inspiration from Monte Carlo sampling to stochastically select diverse expert combinations during training. 
This strategy efficiently increases the co-occurrence frequency of expert combinations without incurring significant training overhead, thereby enabling the model to learn collaborative capabilities required for effective inference with high expert counts. Second, to ensure reliable performance across varying computational budgets, we introduce the \textbf{hierarchical router loss}, which leverages KL divergence to push the router's output distribution away from uniformity, thereby imposing a clear hierarchical ranking upon the experts for each token. This yields a high-quality set of top-\textit{k} experts across budgets, allowing the model to scale gracefully with available computation.

We conduct extensive experiments across LoRA-based and FFN-based MoE scenarios on four model architectures with varying parameter scales, assessed across nine benchmarks. Results show that, unlike standard Top-$k$ models, EMoE achieves monotonically increasing performance as the number of activated experts grows, with the effective scaling range reaching up to 
2-3$\times$ the training-time $k$. Moreover, it consistently outperforms baselines under various computational budgets ($k'$), highlighting its strong utility in diverse settings. Further experimental analysis confirms that both stochastic co-activation sampling and the hierarchical router loss are crucial to EMoE's effectiveness. In summary, these results collectively establish EMoE as a powerful and practical framework that successfully unlocks the elastic potential of MoE models during inference.

\vspace{-0.1cm}
\section{Related Work}
\textbf{Mixture-of-Experts.} The MoE architecture increases model capacity while controlling 
computational costs by activating only a subset of parameters per 
input~\citep{DBLP:journals/neco/JacobsJNH91,DBLP:conf/iclr/ShazeerMMDLHD17,DBLP:conf/iclr/LepikhinLXCFHKS21,DBLP:journals/jmlr/FedusZS22}. 
Subsequent research optimizes expert design~\citep{DBLP:journals/corr/abs-2405-04434,DBLP:journals/corr/abs-2408-10681}, routing 
mechanisms~\citep{DBLP:conf/iclr/PuigcerverRMH24,DBLP:conf/iclr/WangZC25}, and load 
balancing~\citep{DBLP:journals/corr/abs-2408-15664}.
Recently, several studies~\citep{huang-etal-2024-harder,DBLP:conf/emnlp/ZengMG0D24,DBLP:conf/iclr/0001Z0Y25} explore dynamic routing, where the number of activated experts varies per token to allocate more computation to complex tokens and less to simpler ones, all under a fixed computational budget. 
Different from previous studies, our work does not focus on redistributing computation under a fixed budget. Instead, we explore how to ensure and enhance the performance of MoE models when the total computational budget changes during inference. Our goal is to endow MoE models with inference-time scalability.

\textbf{Inference-Time Computational Scaling.} Scaling computation at inference is a strong strategy for addressing the performance-efficiency trade-off in LLMs~\citep{DBLP:journals/corr/abs-2408-03314}. Current studies mainly explore two dimensions: the depth dimension, which aims to enhance model ability by increasing the length of the reasoning chain during inference~\citep{DBLP:conf/acl/ChenSZXSLWSW025,DBLP:journals/corr/abs-2507-10524}; and the width dimension, where prior work primarily on dense models extracts subnetworks of varying sizes from a pre-trained large model by drawing on the concept of pruning~\citep{DBLP:conf/nips/DevvritKKDCDTHK24,DBLP:conf/nips/HabererHL24}, to accommodate different hardware or latency constraints. Our work adopts a fundamentally new perspective on inference-time scalability for MoE models. Instead of increasing model depth or extracting sub-models, we are the first to explore how to effectively utilize increased computational budgets by activating and combining a greater number of experts. This compositional approach to computation scaling at inference allows the model to transition smoothly from a sparse activation state to a denser one, thereby unlocking its full potential in accordance with available resources.
\vspace{-0.1cm}
\section{An Empirical Study on Inference-Time Expert Scaling}
\begin{figure*}[t]
\centering
\subfloat[Trained with 2 activated experts]{\includegraphics[width=0.48\linewidth]{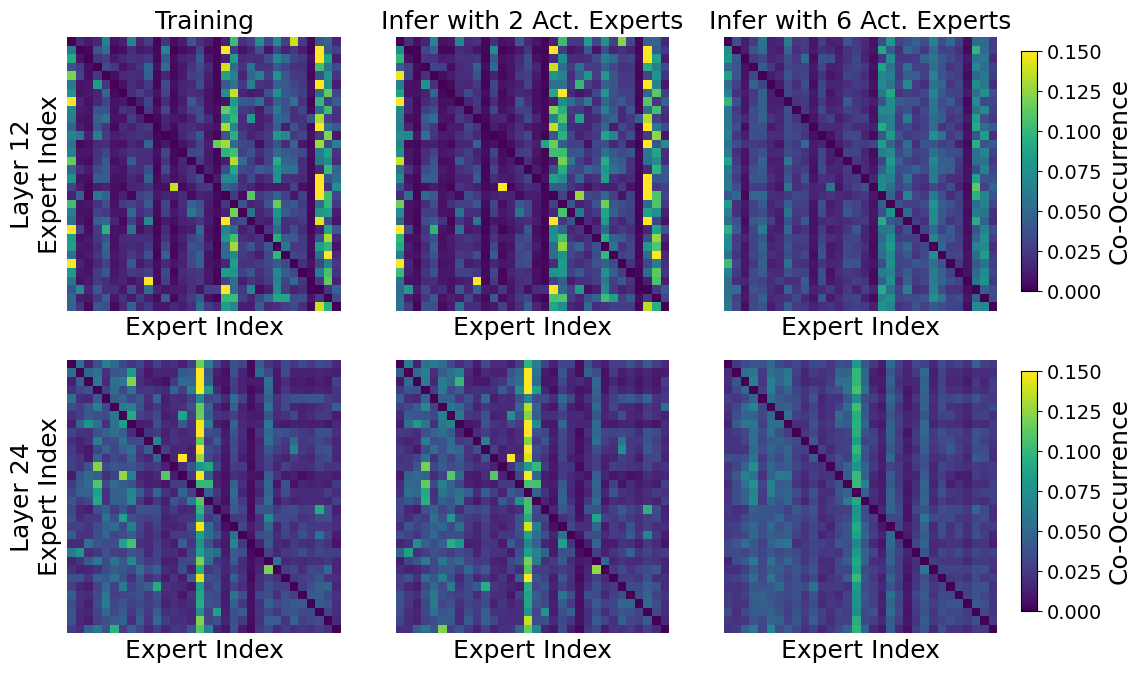}\label{fig:pilot2-k2}} 
\hspace{0.05\linewidth} 
\subfloat[Trained with 6 activated experts]{ \includegraphics[width=0.34\linewidth]{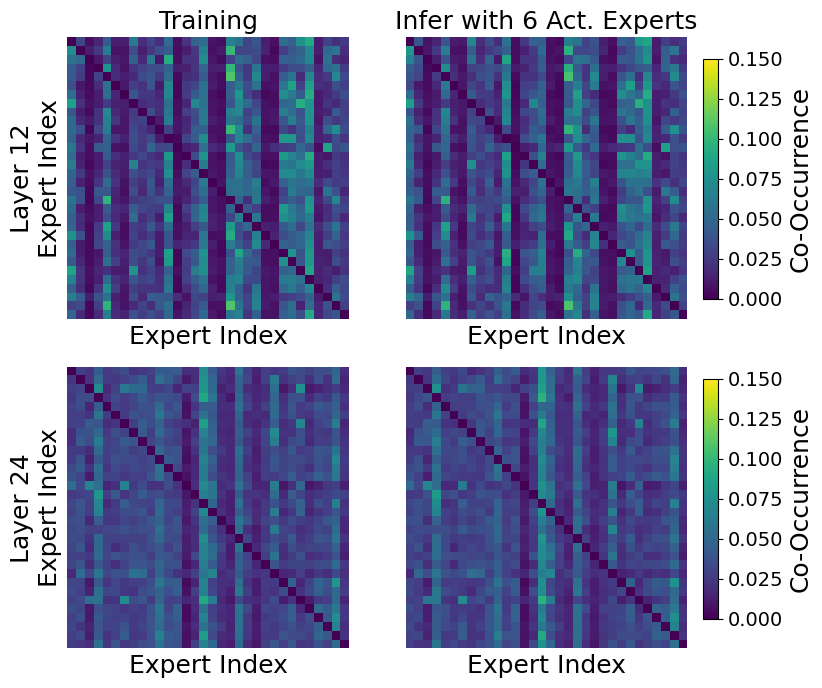}\label{fig:pilot2-k6}}
\caption{Visualization of expert co-occurrence matrices. Panels show models trained with (a) $k=2$ and (b) $k=6$ experts. Each panel compares the co-occurrence patterns observed during training and inference. Extrapolating from $k=2$ to $k'=6$ substantially changes the co-activation structure.}
\label{fig:pilot2}
\end{figure*}
\vspace{-0.1cm}
\subsection{Preliminaries}
A standard MoE layer consists of $N$ experts ${E_i(\cdot; \theta_i)}_{i=1}^N$ and a router $G$ that selects a sparse subset for each input token. Given a token representation $x$, the router computes logits $h(x) = \mathbf{W}_g x$, where $\mathbf{W}_g$ is a learnable matrix. Under Top-$k$ gating, the $k$ experts with the highest logits are selected. Let $\pi(x)$ denote the permutation sorting $h(x)$ in descending order. The active expert set $\mathcal{S}_k(x)$ is:
\begin{equation}
    \mathcal{S}_k(x) = \{\pi_1(x), \pi_2(x), \dots, \pi_k(x)\}.
\end{equation}
The final output $y(x)$ is a weighted sum of active experts, with softmax-normalized logits as weights:
\begin{equation}
    y(x) = \sum_{i \in \mathcal{S}_k(x)}
    \frac{\exp(h_i(x))}{\sum_{j \in \mathcal{S}_k(x)} \exp(h_j(x))} \cdot E_i(x; \theta_i).
\end{equation}
\vspace{-0.1cm}
\subsection{Findings and Analysis}
\label{sec:pilot}
\begin{wrapfigure}[15]{r}{0.40\textwidth}
\vspace{-0.5cm}
  \centering
  \includegraphics[width=0.35\textwidth]{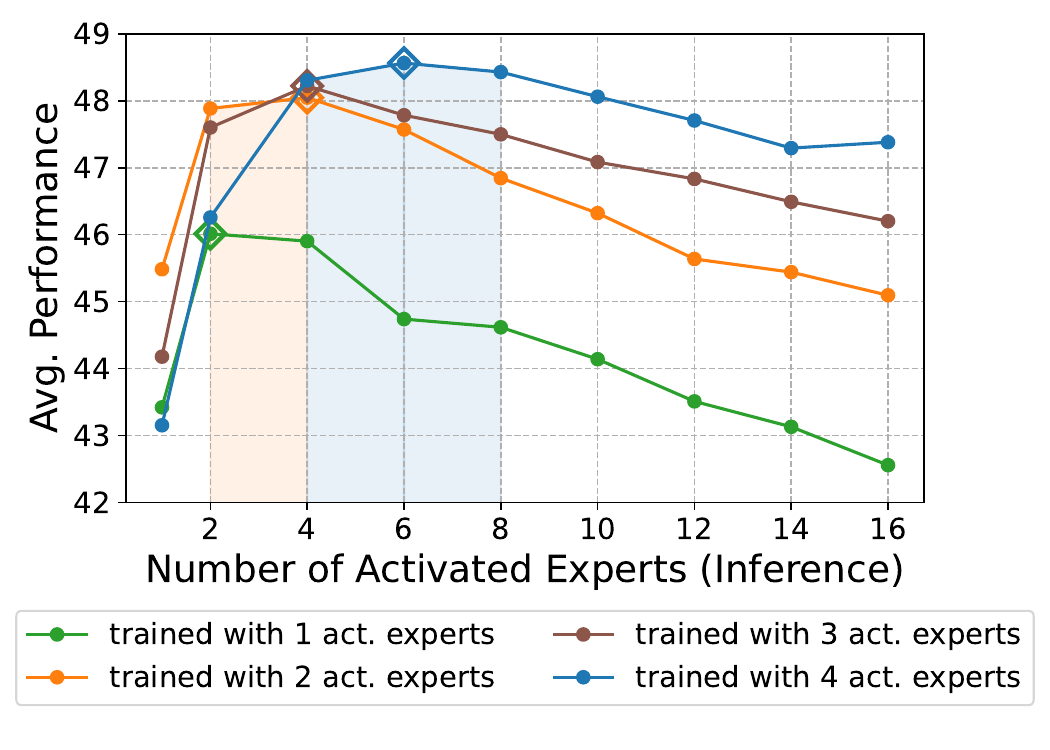}
  \vspace{-0.2cm}
  \caption{Performance of MoE models trained with fixed $k$ under varying inference-time activated experts ($k'$). The color regions show where optimal performance briefly holds.}
  \label{fig:pilot1}
\end{wrapfigure}
Given the practical need for MoE models to operate across 
varying inference budgets, a natural strategy is to activate more experts 
when additional resources are available, thereby leveraging a larger 
portion of the model's total capacity.
To investigate this, we train a LLaMA2-7B model equipped with LoRAMoE~\citep{dou2024loramoealleviateworldknowledge} containing 32 experts. We train separate models, each with a different number of activated experts $k$, and report average performance across nine benchmarks (Appendix~\ref{sec:imp}).

\textbf{Findings.} As shown in Figure~\ref{fig:pilot1}, we identify a counter-intuitive \textbf{inference-time scaling wall}: when the number of experts activated at inference $k'$ exceeds the training budget (e.g., $k=2$), performance holds briefly but quickly drops thereafter, even though more parameters are utilized. While larger training budgets shift this peak, they are computationally prohibitive and suffer from performance collapse when the inference budget is reduced ($k' < k$), leading to performance collapse when returning to the original inference budget. Empirically, this occurs because the model learns to rely solely on activating many experts simultaneously, without teaching the router how to make effective selections under conventional budgets. These findings motivate our central goal: to develop a method that provides elasticity across different inference-time budgets while preserving the standard training cost of conventional $k$ configurations.

\noindent\textbf{Analysis.} To diagnose the cause of the inability of models trained with $k$ to extrapolate to larger $k'$, we investigate the discrepancy in expert activation patterns between training and inference. We introduce an expert co-occurrence matrix, $M^{(k)} \in \mathbb{R}^{N \times N}$, to quantify the frequency with which any two experts are activated together for the same token:
\begin{equation}
\label{eq:co-matrix}
    M_{ij}^{(k)} = \frac{1}{|D|} \sum_{x \in D} \mathbf{1}[i \in \mathcal{S}_{k}(x) \land j \in \mathcal{S}_{k}(x)],
\end{equation}
where $D$ denotes the dataset and $\mathcal{S}_{k}(x)$ the expert set selected by Top-$k$ gating for token $x$. Figures~\ref{fig:pilot2-k2} visualize these co-occurrence matrices for models trained with $k=2$ and $k=6$ experts. For the model trained with $k=2$, the co-occurrence matrix observed during training is sparse, reflecting a specific set of learned expert combinations. However, when it is subjected to inference with $k'=6$, the matrix becomes much denser and qualitatively different. This indicates that the model is forced to utilize many expert combinations that are seldom, if ever, seen during training. These experts have not been optimized to collaborate, leading to a breakdown in their collective output. Conversely, for the model trained with $k=6$ (Figures~\ref{fig:pilot2-k6}), the co-occurrence matrices from training and inference are structurally similar. This alignment between training and inference conditions explains why the model's performance is more stable. We therefore hypothesize that the inability of MoE models to extrapolate to higher expert counts stems from a lack of collaborative training among the sparsely activated experts.

\begin{wrapfigure}[13]{r}{0.35\textwidth}
\vspace{-0.45cm}
  \centering
  \includegraphics[width=0.35\textwidth]{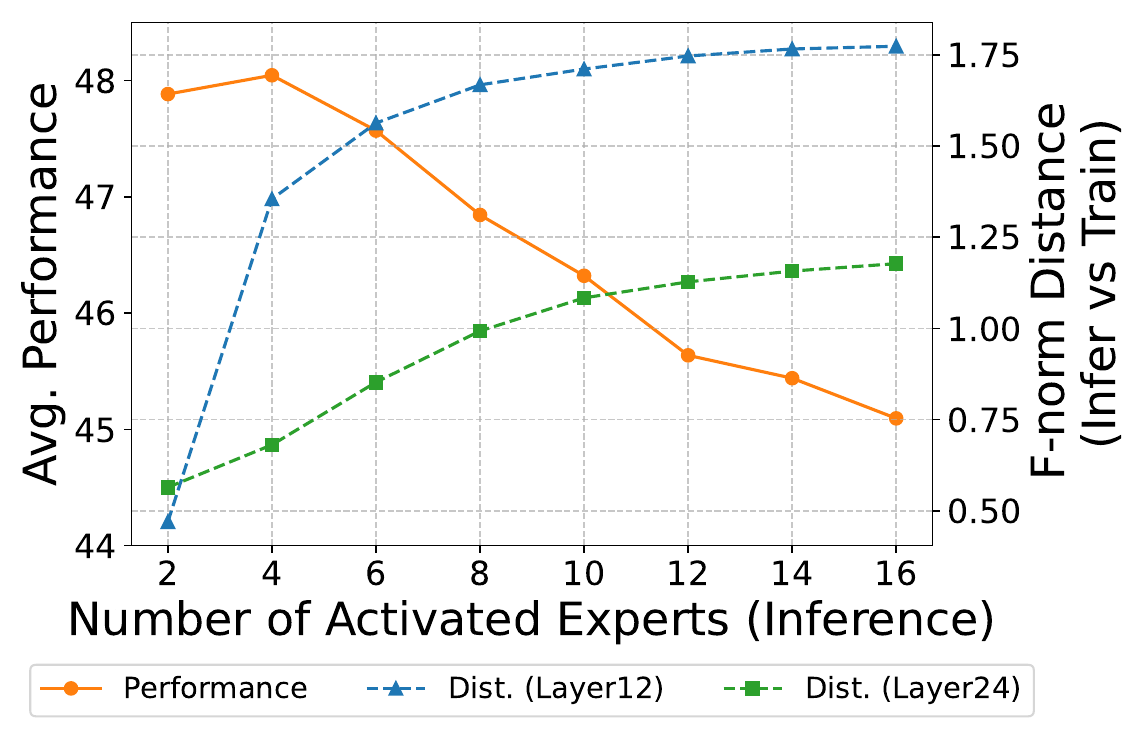}
  \vspace{-0.6cm}
  \caption{Co-occurrence distance vs. model performance for a model trained with $k=2$.}
  \label{fig:pilot3}
\end{wrapfigure}
To quantify the impact of discrepancy, we measure the Frobenius norm of the distance between the co-occurrence matrix from training $M^{(k)}$ and inference $M^{(k')}$:
\begin{equation}
    \Delta(k \to k') = \| M^{(k)} - M^{(k')} \|_F.
\end{equation}
This metric captures the distance in expert activation patterns. A small $\Delta$ indicates that the expert combinations encountered at inference are similar to the distribution seen during training. Conversely, a large $\Delta$ signifies a severe distributional shift, where the model is used on untested expert collaborations. Figure~\ref{fig:pilot3} plots this relationship for a model trained with $k=2$ experts. The results show a clear and compelling trend. As $k'$ increases, the F-norm distance $\Delta$ grows monotonically, which is anti-correlated with model performance. Beyond the optimal point, every subsequent increase in the number of experts leads to a larger co-occurrence distance and a corresponding, significant drop in performance. This evidence suggests that the model’s reliance on new expert combinations that have not been sufficiently trained to collaborate is a key factor behind the scaling wall.

\vspace{-0.1cm}
\section{Elastic Mixture-of-Experts}
To enable elastic inference across varying computational budgets, we introduce Elastic Mixture-of-Experts (EMoE) as shown in Figure~\ref{fig:method}. EMoE incorporates two designs: stochastic co-activation sampling, which resolves the expert co-occurrence discrepancy by training diverse combinations of experts to collaborate effectively, and a hierarchical router loss, which encourages stable and decisive expert rankings. Together, these ensure robust performance across varying computational budgets.
\begin{figure*}[t]
    \centering
\includegraphics[width=1.0\columnwidth,trim=0 30 0 0,clip]{figs/EMoE-v1.5.pdf}
\caption{Comparison of the standard Top-$k$ MoE and our Elastic Mixture-of-Experts (EMoE). EMoE is designed to unlock scalability at inference time. For each input, it first forms a candidate pool $\mathcal{S}_{k_{\text{ideal}}}$ of top-scoring experts. A smaller subset $\mathcal{S}_{\text{co-act}}$ is then uniformly drawn from this pool for computation. The total objective combines standard MoE losses with the hierarchical router loss $\mathcal{L}_\text{HR}$, which encourages the router to produce a decisive, non-uniform expert distribution.}
    \label{fig:method}
\end{figure*}
\subsection{Stochastic Co-activation Sampling}
Our analysis reveals that the extrapolation failure stems from insufficient learning of diverse expert combinations during training. While training with a larger $k$ can alleviate this by covering more combinations, it incurs prohibitive computational overhead and loses elasticity at lower inference budgets (as shown in Figure~\ref{fig:pilot1}). We seek an alternative: can we expose the model to diverse expert 
combinations while retaining the efficiency of standard-$k$ training? Inspired by Monte Carlo sampling, we propose stochastic co-activation sampling. The key insight is to decouple the \emph{training budget} (determined by $k_\text{train}$) from the \emph{combination coverage} (determined by $k_\text{ideal}$). Specifically, for each token $x$, we first 
identify a larger candidate pool $\mathcal{S}_{k_\text{ideal}}(x)$ 
from the router's top-ranked experts, then sample a subset of 
size $k_\text{train}$ for actual computation:
\begin{equation}
\mathcal{S}_\text{co-act}(x) \sim \text{UniformSample}(\mathcal{S}_{k_\text{ideal}}(x), k_\text{train}),
\label{eq:sample}
\end{equation}
and compute the MoE output $y_\text{co-act}(x)$ on this subset:
\begin{equation}
y_\text{co-act}(x) = \sum_{i \in \mathcal{S}_\text{co-act}(x)}
\frac{\exp(h_i(x))}{\sum_{j \in \mathcal{S}_\text{co-act}(x)} \exp(h_j(x))} \cdot E_i(x;\theta_i).
\end{equation}
Over multiple training steps, this provides a stochastic approximation that captures diverse expert combinations.

To further ease the optimization burden, we introduce a dynamic sampling process in practice. This strategy replaces the fixed-size $k_\text{ideal}$ with a variable one, adjusting the sampling space to stabilize training. For each input token, we stochastically determine the size of a candidate pool $\tilde{k}_\text{ideal}$, by drawing it uniformly from the integer interval $[k_\text{train}, k_\text{ideal}]$. It ensures that the candidate pool is frequently drawn from a smaller, higher-confidence set (i.e., when $\tilde{k}_\text{ideal}$ is sampled to be close to $k_\text{train}$) and guarantees that the core group of top experts receives consistent and focused training signals. Concurrently, the uniform sampling up to $k_\text{ideal}$ introduces controlled exploration, allowing the model to learn diverse co-activation patterns. From this dynamically sized candidate pool, we then perform the above sampling step in Eq~\ref{eq:sample}, selecting a final training subset $\mathcal{S}_\text{co-act}(x)$.

\textbf{Why Co-activation Sampling Works?} The efficacy can be directly understood by examining its impact on the expert co-occurrence matrix $M^{(k)}$ from our pilot study. We previously established that performance degradation when scaling from a training budget $k$ to an inference budget $k'$ correlates strongly with a large co-occurrence distance $\Delta(k \to k') = \| M^{(k)} - M^{(k')} \|_F$. This distance arises because many entries in the matrix $M_{ij}^{(k)}$ during training are zero or near-zero, while the corresponding entries $M_{ij}^{(k')}$ become substantially non-zero at inference, forcing the model to rely on untested expert combinations.

Co-activation sampling is designed to minimize this future discrepancy by ``filling in'' the sparse training co-occurrence matrix in advance. The mechanism is probabilistic. For any token where two experts $i$ and $j$ fall within the candidate pool $\mathcal{S}_{k_{\text{ideal}}}(x)$, their probability of being jointly selected for a training update is uniformly defined as:
\begin{equation}
   P(i, j \in \mathcal{S}_\text{co-act}(x) \mid i, j \in  \mathcal{S}_{k_{\text{ideal}}}(x)) = \frac{C({k_{\text{ideal}}-2},{k_{\text{train}}-2})}{C({k_{\text{ideal}}},{k_{\text{train}}})}.
    \label{eq:sct_prob_link} 
\end{equation}
This ensures that a wide range of expert pairs receive collaborative training signals. Let's revisit our concrete example ($N=32$ experts, standard training $k=2$, versus EMoE with $k_{\text{train}}=2, k_{\text{ideal}}=8$). Consider an expert pair $(i, j)$ where one or both experts are ranked outside the top-2 but within the top-8. In standard training, their co-occurrence entry $M_{ij}^{(2)}$ is effectively zero. With co-activation sampling, assuming this pair is in the top-8 pool for a given token, their co-activation probability becomes $C({6},{0}) / C({8},{2}) = 1/28 \approx 3.6\%$. While this probability seems small for a single instance, when aggregated over multiple training steps, it guarantees that the corresponding entry $(M_\text{co-act}^{(2)})_{ij}$ becomes substantially non-zero, mitigating the distance with the co-occurrence matrix at inference.
\vspace{-0.1cm}
\subsection{Hierarchical Router Loss}
While co-activation sampling addresses the challenge of scaling \emph{up}, a complementary problem remains: the router does not reliably make effective selections under low inference budgets, leading to degraded performance (Figure~\ref{fig:pilot1}). Since the inference budget 
$k'$ can vary in practical deployments, we expect the model to achieve strong performance across different inference budget. A key issue arises when the router assigns nearly uniform weights to experts: in this case, the distinction between Top-$k$ and the rest becomes ambiguous, and a small $k'$ will underperform. Therefore, the router should produce a clear, hierarchical expert ranking for each token, such that activating only a few experts (small $k'$) or many experts (large $k'$) both lead to reliable performance. To achieve this, we encourage the router distribution $q(x)=\mathrm{softmax}(h(x))$ to be far from a uniform distribution. Concretely, we introduce a KL-based regularization:
\begin{equation}
    \mathcal{L}_\text{HR} = -D_{KL}\!\left(q(x) \,\|\, \mathcal{U}\right) 
    = -\sum_{i=1}^{N} q_i(x) \log\!\left(\frac{q_i(x)}{1/N}\right),
\end{equation}
where $\mathcal{U}$ denotes the uniform distribution over experts. Here we use reverse KL rather than forward KL. Using forward KL, i.e., $-D_{KL}(\mathcal{U}\,\|\,q(x)) = \frac{1}{N}\sum_i (\log q_i(x) + \log N)$, the resulting gradients would be $-\frac{\partial}{\partial q_i} D_{KL}(\mathcal{U} \,\|\, q(x)) = \frac{1}{N q_i(x)}$. In contrast, reverse KL yields smoother gradients $\partial \mathcal{L}_\text{HR}/\partial q_i = -\log(q_i(x) N) - 1$. The gradients of forward KL increase more rapidly as $q_i(x)$ approaches zero, since $1/q_i(x)$ diverges much faster than $-\log q_i(x)$ does. For example, when $q_i(x) = 0.01$, we have $1/q_i(x) = 100$, while $-\log q_i(x) \approx 4.6$. This sharp increase in gradients can cause instability during training. By contrast, the reverse KL sharpens the distribution without excessive concentration, preserving stable Top-$k$ rankings while maintaining the potential contribution of other experts. Thus, the full EMoE training objective augments the standard MoE loss $\mathcal{L}_\text{MoE}$ (comprising cross-entropy $\mathcal{L}_{ce}$ and load balancing $\mathcal{L}_{b}$) with our proposed loss: $\mathcal{L} = \mathcal{L}_\text{MoE} + \lambda \cdot \mathcal{L}_\text{HR}$, where $\lambda$ is a balancing coefficient. Unlike the load balance loss $\mathcal{L}_{b}$, which ensures even utilization of experts across the dataset, $\mathcal{L}_\text{HR}$ operates at the token level, producing a decisive ranking for each individual routing decision. Appendix~\ref{app:lbl} provides an empirical analysis confirming their complementary effects.

\begin{table*}[t]
\caption{Comparison between EMoE and Top-$k$ across inference budgets 
$k'$. For DeepSeek-V2-Lite and 
ERNIE-4.5-21B, ``$6+2$'' denotes 6 routed + 2 shared experts following their original settings. Both methods are trained with the same budget. Results 
are averaged over three runs with standard deviation. Appendix~\ref{app:large-k} and~\ref{app:ktrain-scaling} evaluate EMoE against varied training budgets and large-$k$ configurations.}
\small
\centering
\setlength{\tabcolsep}{1.5pt}
\scalebox{0.95}{
\begin{tabular}{@{}lccccccccc>{\columncolor{gray!20}}cc@{}@{}}
\toprule
& \textbf{ARC-c} & \textbf{ARC-e} & \textbf{GSM8K} & \textbf{HellaS.} & \textbf{HumanE.} & \textbf{NQ} & \textbf{Tri.QA} & \textbf{Wino.} & \textbf{MMLU} & \textbf{AVG} \\ \midrule
\midrule
\multicolumn{11}{l}{\textit{LoRAMoE (trained with 2 activated experts)}} \\
Top-$k$ ($k' = 1$) & 56.95 & 73.55 & 33.33 & 52.71 & 15.97 & 22.89 & 54.32 & 53.59 & 46.04 & 45.48$_{\pm\text{0.91}}$ \\
Top-$k$ ($k' = 2$) & 56.54 & 75.98 & 37.54 & 54.87 & 20.24 & 25.55 & 57.62 & 55.06 & 47.55 & 47.88$_{\pm\text{0.57}}$ \\
Top-$k$ ($k' = 4$) & 56.75 & 76.65 & 38.16 & 53.07 & 19.39 & 26.91 & 59.06 & 55.19 & 47.23 & 48.05$_{\pm\text{0.36}}$ \\
Top-$k$ ($k' = 6$) & 57.74 & 75.60 & 35.94 & 50.76 & 18.70 & 27.17 & 59.53 & 55.49 & 47.20 & 47.57$_{\pm\text{0.52}}$ \\
\midrule
EMoE ($k' = 1$) & 55.25 & 71.78 & 31.24 & 51.52 & 17.68 & 25.46 & 54.79 & 56.75 & 46.64 & \textbf{45.68}$_{\pm\text{1.19}}$ \\
EMoE ($k' = 2$) & 56.95 & 78.66 & 37.98 & 53.57 & 18.90 & 26.62 & 58.20 & 55.41 & 47.65 & \textbf{48.22}$_{\pm\text{0.62}}$ \\
EMoE ($k' = 4$) & 58.31 & 79.72 & 38.21 & 55.95 & 18.29 & 27.78 & 59.24 & 55.64 & 47.89 & \textbf{49.00}$_{\pm\text{0.70}}$ \\
EMoE ($k' = 6$) & 60.34 & 79.37 & 38.21 & 56.14 & 20.73 & 27.87 & 59.77 & 55.33 & 47.73 & \textbf{49.50}$_{\pm\text{0.65}}$ \\
\midrule
\midrule
\multicolumn{11}{l}{\textit{
OLMoE-1B-7B-0924 (trained with 8 activated experts)}} \\
Top-$k$ ($k' = 4$) & 43.73 & 65.96 & 17.13 & 41.40 & 13.41 & 14.57 & 32.04 & 50.36 & 39.17 & 35.31$_{\pm\text{1.06}}$ \\
Top-$k$ ($k' = 8$) & 52.54 & 71.78 & 24.94 & 48.68 & 21.34 & 17.95 & 38.98 & 52.09 & 43.85 & 41.35$_{\pm\text{0.55}}$ \\
Top-$k$ ($k' = 16$) & 53.56 & 72.31 & 25.85 & 48.59 & 17.07 & 18.12 & 39.10 & 51.62 & 43.61 & 41.09$_{\pm\text{0.75}}$ \\
\midrule
EMoE ($k' = 4$) & 44.75 & 73.54 & 22.21 & 43.24 & 15.85 & 14.52 & 34.84 & 51.85 & 39.45 & \textbf{37.81}$_{\pm\text{0.72}}$ \\
EMoE ($k' = 8$) & 52.20 & 75.13 & 26.46 & 51.22 & 20.12 & 18.67 & 39.64 & 52.72 & 41.56 & \textbf{41.97}$_{\pm\text{0.59}}$ \\
EMoE ($k' = 16$) & 55.93 & 76.37 & 27.75 & 50.69 & 21.95 & 19.31 & 41.27 & 52.09 & 42.79 & \textbf{43.13}$_{\pm\text{0.74}}$ \\
\midrule
\midrule
\multicolumn{11}{l}{\textit{DeepSeek-V2-Lite (trained with 6+2 activated experts)}} \\
Top-$k$ ($k' = 3+2$) & 58.98 & 72.66 & 42.99 & 57.36 & 28.66 & 18.89 & 41.36 & 55.96 & 47.76 & 47.18$_{\pm\text{0.48}}$ \\
Top-$k$ ($k' = 6+2$) & 64.07 & 75.49 & 49.36 & 58.52 & 36.59 & 21.63 & 46.86 & 57.22 & 49.88 & 51.07$_{\pm\text{0.01}}$ \\
Top-$k$ ($k' = 12+2$) & 62.71 & 74.43 & 50.19 & 57.60 & 39.02 & 21.05 & 46.38 & 57.22 & 49.89 & 50.94$_{\pm\text{0.16}}$ \\
\midrule
EMoE ($k' = 3+2$) & 59.66 & 78.66 & 45.34 & 59.61 & 34.76 & 20.44 & 47.81 & 56.35 & 48.52 & \textbf{50.13}$_{\pm\text{0.64}}$ \\
EMoE ($k' = 6+2$) & 63.73 & 82.54 & 48.52 & 61.29 & 39.63 & 21.72 & 51.58 & 56.67 & 49.30 & \textbf{52.78}$_{\pm\text{0.18}}$ \\
EMoE ($k' = 12+2$) & 65.42 & 83.95 & 50.87 & 62.25 & 41.46 & 22.80 & 52.07 & 56.99 & 51.94 & \textbf{54.19}$_{\pm\text{0.33}}$ \\
\midrule
\midrule
\multicolumn{11}{l}{\textit{ERNIE-4.5-21B-A3B (trained with 6+2 activated experts)}} \\
Top-$k$ ($k' = 3+2$) & 88.81 & 94.18 & 75.44 & 80.86 & 66.46 & 26.18 & 59.10 & 64.17 & 71.58 & 69.64$_{\pm\text{0.15}}$ \\
Top-$k$ ($k' = 6+2$) & 89.15 & 95.24 & 79.83 & 81.56 & 71.34 & 26.43 & 61.10 & 66.93 & 72.86 & 71.60$_{\pm\text{0.23}}$ \\
Top-$k$ ($k' = 12+2$) & 88.81 & 94.89 & 80.52 & 80.08 & 68.29 & 26.81 & 60.24 & 67.09 & 71.92 & 70.96$_{\pm\text{0.41}}$ \\
\midrule
EMoE ($k' = 3+2$) & 88.47 & 94.18 & 75.51 & 82.96 & 70.73 & 25.37 & 59.22 & 65.35 & 72.28 & \textbf{70.45}$_{\pm\text{0.12}}$ \\
EMoE ($k' = 6+2$) & 88.14 & 94.71 & 77.71 & 84.18 & 73.17 & 25.62 & 61.32 & 68.27 & 72.65 & \textbf{71.75}$_{\pm\text{0.26}}$ \\
EMoE ($k' = 12+2$) & 90.85 & 95.94 & 81.73 & 84.54 & 75.61 & 26.15 & 60.13 & 68.51 & 72.93 & \textbf{72.93}$_{\pm\text{0.55}}$ \\ \bottomrule
\end{tabular}
}
\label{tab:main1}
\end{table*}
\textbf{Putting them together.} EMoE provides a lightweight yet powerful framework for training inference-scalable MoE models. Stochastic co-activation sampling directly tackles the problem of collaboration failure by teaching experts to collaborate within diverse, stochastically sampled combinations. Concurrently, the hierarchical loss guides the router to learn a stable and decisive expert ranking. Together, they ensure that the model can gracefully and effectively scale its performance to match the given computational budget at inference time, eliminating the need to train or deploy multiple MoE variants tailored to different computational settings. Notably, EMoE maintains the same training cost $k_\text{train}$ as the Top-$k$ method and can be applied during post-training on pretrained MoE checkpoints, without restarting pretraining. A detailed 
analysis of training cost is provided in Appendix~\ref{app:training-eff}.
\vspace{-0.1cm}
\section{Experiments}
\vspace{-0.1cm}
\subsection{Experimental Setup}
\textbf{Model Settings.} Our experiments consider two MoE scenarios: LoRA-based and FFN-based settings. In the LoRA-based scenario, we adopt LLaMA2-7B~\citep{DBLP:journals/corr/abs-2307-09288} as the base model and configure 32 LoRA experts in each layer. In the FFN-based scenario, we evaluate three advanced MoE models of different scales: OLMoE-1B-7B-0924~\citep{muennighoff2024olmoeopenmixtureofexpertslanguage}, DeepSeek-V2-Lite~\citep{DBLP:journals/corr/abs-2405-04434}, and ERNIE-4.5-21B-A3B~\citep{ernie2025technicalreport}.

\textbf{Baselines.} We compare against two categories of baselines. The first category consists of mainstream MoE models that employ a fixed Top-$k$ strategy. The second category includes dynamic routing methods: AdaMoE~\citep{DBLP:conf/emnlp/ZengMG0D24} and Top-$p$~\citep{huang-etal-2024-harder}, which dynamically adjust the number of activated experts across tokens while keeping the total number of activated experts fixed. In contrast, our method allows the total number of activated experts to be flexibly adjusted according to computational budgets.
\begin{table*}[t]
\caption{Comparisons between EMoE and dynamic routing methods across inference budget $k'$. All methods are trained with a training budget equivalent to that of a standard Top-$k$ MoE with $k_\text{train}=2$. For AdaMoE and Top-$p$, $k'$ refers to the average number of activated experts across all tokens.}
\small
\centering
\setlength{\tabcolsep}{2.6pt}
\begin{tabular}{@{}ccccccccccc>{\columncolor{gray!20}}cc@{}}
\toprule
\multicolumn{1}{l}{} & \textbf{$k'$} & \textbf{ARC-c} & \textbf{ARC-e} & \textbf{GSM8K} & \textbf{HellaS.} & \textbf{HumanE.} & \textbf{NQ} & \textbf{Tri.QA} & \textbf{Wino} & \textbf{MMLU} & \textbf{AVG} \\
\midrule
\midrule
\multirow{4}{*}{Top-$k$} & 1.0 & 56.95 & 73.55 & 33.33 & 52.71 & 15.97 & 22.89 & 54.32 & 53.59 & 46.04 & 45.48$_{\pm\text{0.91}}$ \\
 & 2.0 & 56.54 & 75.98 & 37.54 & 54.87 & 20.24 & 25.55 & 57.62 & 55.06 & 47.55 & 47.88$_{\pm\text{0.57}}$ \\
 & 4.0 & 56.75 & 76.65 & 38.16 & 53.07 & 19.39 & 26.91 & 59.06 & 55.19 & 47.23 & 48.05$_{\pm\text{0.36}}$ \\
 & 6.0 & 57.74 & 75.60 & 35.94 & 50.76 & 18.70 & 27.17 & 59.53 & 55.49 & 47.20 & 47.57$_{\pm\text{0.52}}$ \\
 \midrule
\multirow{4}{*}{Top-$p$} & 1.0 & 56.95 & 74.25 & 35.56 & 49.19 & 15.85 & 24.60 & 55.58 & 56.43 & 46.41 & \textbf{46.09}$_{\pm\text{0.33}}$ \\
 & 2.2 & 55.59 & 77.60 & 36.62 & 50.14 & 17.07 & 25.79 & 57.67 & 55.88 & 47.04 & 47.04$_{\pm\text{0.72}}$ \\
 & 4.1 & 59.66 & 79.01 & 36.92 & 49.54 & 20.12 & 27.06 & 59.41 & 54.30 & 46.88 & 48.10$_{\pm\text{1.23}}$ \\
 & 6.0 & 56.95 & 78.84 & 36.92 & 46.48 & 20.12 & 27.12 & 60.14 & 53.51 & 47.25 & 47.48$_{\pm\text{0.69}}$ \\
 \midrule
\multirow{4}{*}{AdaMoE} & 1.3 & 55.93 & 67.20 & 34.12 & 49.83 & 18.29 & 14.74 & 35.95 & 52.17 & 43.83 & 41.34$_{\pm\text{1.42}}$ \\
 & 2.2 & 60.68 & 75.13 & 37.30 & 55.28 & 20.73 & 22.58 & 52.03 & 53.99 & 46.89 & 47.18$_{\pm\text{0.72}}$ \\
 & 4.2 & 58.31 & 77.25 & 37.68 & 56.30 & 20.12 & 27.81 & 58.63 & 54.70 & 46.06 & 48.54$_{\pm\text{0.31}}$ \\
 & 6.1 & 56.95 & 77.07 & 37.60 & 56.11 & 20.73 & 28.95 & 59.25 & 54.30 & 45.76 & 48.52$_{\pm\text{0.40}}$ \\
 \midrule
\multirow{4}{*}{EMoE} & 1.0 & 55.25 & 71.78 & 31.24 & 51.52 & 17.68 & 25.46 & 54.79 & 56.75 & 46.64 & 45.68$_{\pm\text{1.19}}$ \\
 & 2.0 & 56.95 & 78.66 & 37.98 & 53.57 & 18.90 & 26.62 & 58.20 & 55.41 & 47.65 & \textbf{48.22}$_{\pm\text{0.62}}$ \\
 & 4.0 & 58.31 & 79.72 & 38.21 & 55.95 & 18.29 & 27.78 & 59.24 & 55.64 & 47.89 & \textbf{49.00}$_{\pm\text{0.70}}$ \\
 & 6.0 & 60.34 & 79.37 & 38.21 & 56.14 & 20.73 & 27.87 & 59.77 & 55.33 & 47.73 & \textbf{49.50}$_{\pm\text{0.65}}$\\
 \bottomrule
\end{tabular}
\label{tab:main2}
\end{table*}

\textbf{Training and Evaluation Data.} Following~\citet{DBLP:journals/corr/abs-2410-01610}, we construct a diverse instruction-tuning dataset comprising 50K samples spanning three domains: coding, mathematics, and general abilities. Specifically, the dataset incorporates Magicoder~\citep{DBLP:journals/corr/abs-2312-02120} for coding, MetaMathQA~\citep{DBLP:conf/iclr/YuJSYLZKLWL24} for mathematics, and SlimORCA~\citep{SlimOrca} for general abilities. Appendix~\ref{app:larger-data} provides experiments on 100K and 200K training dataset sizes. For evaluation, we assess model performance across a comprehensive suite of nine downstream benchmark datasets, covering knowledge, reasoning, coding, and open-domain QA. 

\textbf{Implementation Details.} In the LoRA-based MoE scenario, we activate 2 experts per layer during training to align with the sparse activation pattern typically used in large-scale models. In the FFN-based scenario, we follow the original pretraining configurations: OLMoE activates 8 experts per layer, while DeepSeek-V2-Lite and ERNIE-4.5-21B activate 6 fine-grained experts and 2 shared experts per layer. All MoE models are trained for 4 epochs. The learning rate is set to $2 \times 10^{-4}$ for LoRA-based settings and $2 \times 10^{-5}$ for FFN-based settings. All experiments are conducted three times, and we report the average results along with the standard deviation. More comprehensive details about baselines, data, and implementations are provided in Appendix~\ref{sec:imp}.

\subsection{Main Results}
\label{sec:ana}
\textbf{Comparisons to the Top-$k$ Method on Different Models.} Table~\ref{tab:main1} evaluates EMoE against standard Top-$k$ across four model architectures. Consistent with Section~\ref{sec:pilot}, Top-$k$ models degrade when the number of activated experts exceeds the training budget.
In contrast, models trained with the EMoE framework exhibit robust and monotonically increasing performance scalability. For every model architecture, increasing the number of activated experts at inference consistently leads to performance gains, confirming the effectiveness of the co-activation sampling. Notably, EMoE not only eliminates the performance drop observed in baselines but also leverages the proposed hierarchical loss to deliver further improvements under varying computational budgets, ultimately reaching new peaks in performance. Appendix~\ref{app:large-k} and~\ref{app:ktrain-scaling} further evaluate EMoE against standard large-$k$ and varied training budgets configurations.

\textbf{Comparisons to Dynamic Routing Methods.} In Table~\ref{tab:main2}, we further analyze EMoE and compare it with dynamic routing strategies. These methods are designed to optimize computational resource allocation under a fixed global computation budget by reallocating experts from simpler tokens to more complex ones. The results show that although these dynamic methods do provide some improvements over the static Top-$k$ baseline, 
they ultimately still face the same issue of performance degradation. Their performance either plateaus or begins to degrade after reaching a peak, because these approaches are fundamentally not designed to go beyond a fixed computational limit. In contrast, EMoE demonstrates a distinctly superior scaling trend. Its performance increases monotonically as the number of activated experts grows.
This highlights a key distinction: prior methods focus on optimal reallocation under a fixed compute budget, whereas EMoE is uniquely designed to efficiently utilize variable and scalable computational resources.

\subsection{Analysis}
\begin{wraptable}[11]{ht}{.54\textwidth}
    \small
\vspace{-4.5mm}
\caption{Ablation study on EMoE’s two designs: stochastic co-activation sampling (co-act.) and the hierarchical router loss ($\mathcal{L}_\text{HR}$).}
\small
\setlength{\tabcolsep}{2.9mm}{
\begin{tabular}{@{}lcccc@{}}
\toprule
 & $k'=1$ & $k'=2$ & $k'=4$ & $k'=6$ \\ \midrule
Top-${k}$ & 45.48 & 47.88 & 48.05 & 47.57 \\
\midrule
EMoE & 45.68 & 48.22 & 49.00 & 49.50 \\
\ \ \ w/o co-act. & 45.83 & 48.03 & 48.68 & 48.15 \\ 
\ \ \ w/o $\mathcal{L}_\text{HR}$ & 45.19 & 47.79 & 48.81 & 49.08 \\
\bottomrule
\end{tabular}
}
\label{tab:ablation}
\end{wraptable} 
\textbf{Ablation Study.} Table~\ref{tab:ablation} presents the individual contributions of EMoE's two key designs: stochastic co-activation sampling and hierarchical router loss ($\mathcal{L}_{\text{HR}}$). The variant without co-activation sampling performs well at $k'=1$ due to $\mathcal{L}_{\text{HR}}$'s hierarchical ranking, but performance drops sharply at $k'=6$, indicating a collaborative failure. Conversely, the variant without $\mathcal{L}_{\text{HR}}$ maintains robust performance at $k'=6$ but consistently underperforms the full EMoE model, especially at $k'=1$. These results underscore that both designs are essential: co-activation sampling fosters expert collaboration for scaling, while $\mathcal{L}_{\text{HR}}$ ensures a stable ranking across budgets. Only their combination fully realizes EMoE's potential for inference-time scalability.

\begin{wrapfigure}[13]{r}{0.37\textwidth}
\vspace{-0.6cm}
  \includegraphics[width=0.37\textwidth]{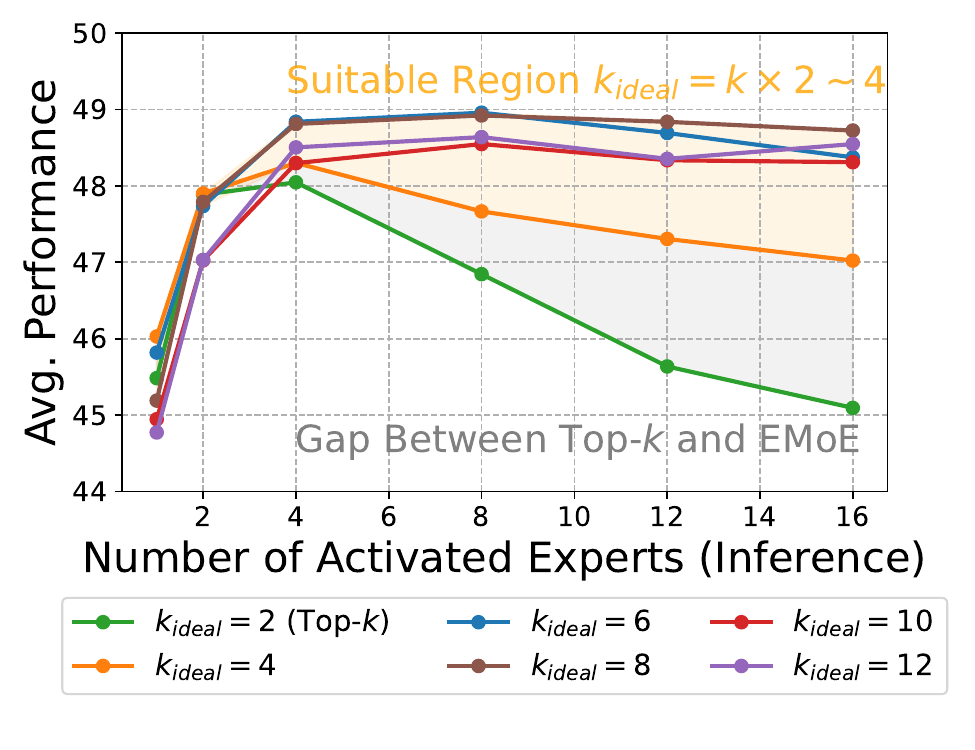}
  \vspace{-0.7cm}
  \caption{
  Analysis of the effect of the hyperparameter $k_\text{ideal}$. All experiments are conducted with $k_\text{train}=2$.}
  \label{fig:ana-2}
\end{wrapfigure}
\textbf{Effect of $k_{\text{ideal}}$.} We analyze the robustness of $k_\text{ideal}$ in Figure~\ref{fig:ana-2}. The choice is highly flexible, as even $k_\text{ideal} = 2 \times k_\text{train}$ already yields significant gains over standard Top-$k$, with optimal performance achieved within $2\text{-}4\times k_\text{train}$. When $k_\text{ideal}$ is set too high (e.g., exceeding $4 \times k_\text{train}=8$), a trade-off emerges: while the model maintains strong performance under high inference budgets, its performance with a low number of activated experts, as well as its overall peak performance, begin to degrade. Analysis on DeepSeek-V2-Lite further confirms the validity of this relaxed range, as shown in Appendix~\ref{app:ideal-dsv2}. Based on this analysis, we  choose $k_\text{ideal} \in\{2,3,4\}$, using $4 \times k_\text{train}$ for the LoRA-based models, $3 \times k_\text{train}$ for DeepSeek-V2-Lite and $2 \times k_\text{train}$ for OLMoE.
\begin{figure*}[t]
\centering
\subfloat[Standard Top-$k$ MoE]{\includegraphics[width=0.33\linewidth]{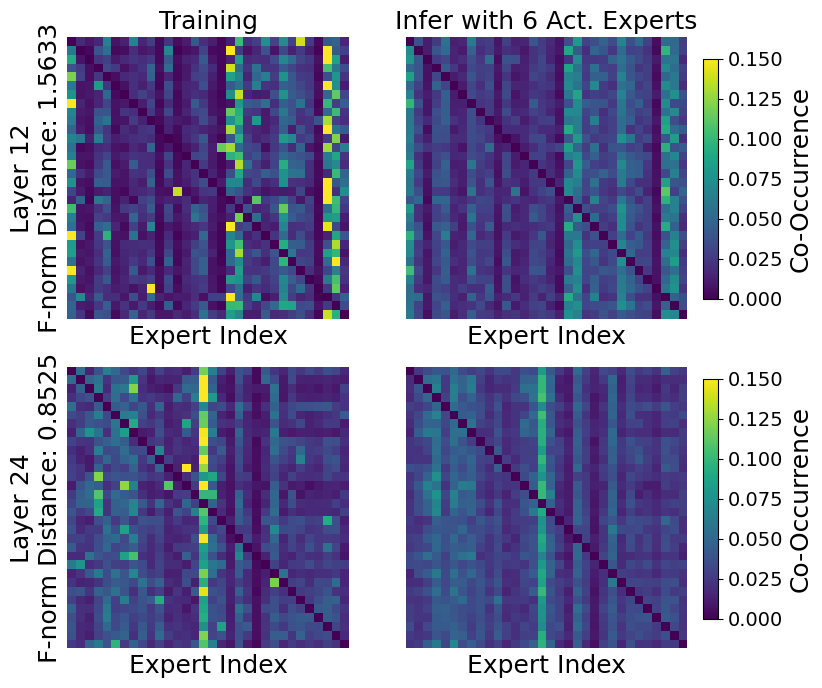}\label{fig:ana3-topk}}
\hspace{0.05\linewidth} 
\subfloat[Our proposed Elastic MoE]{ \includegraphics[width=0.33\linewidth]{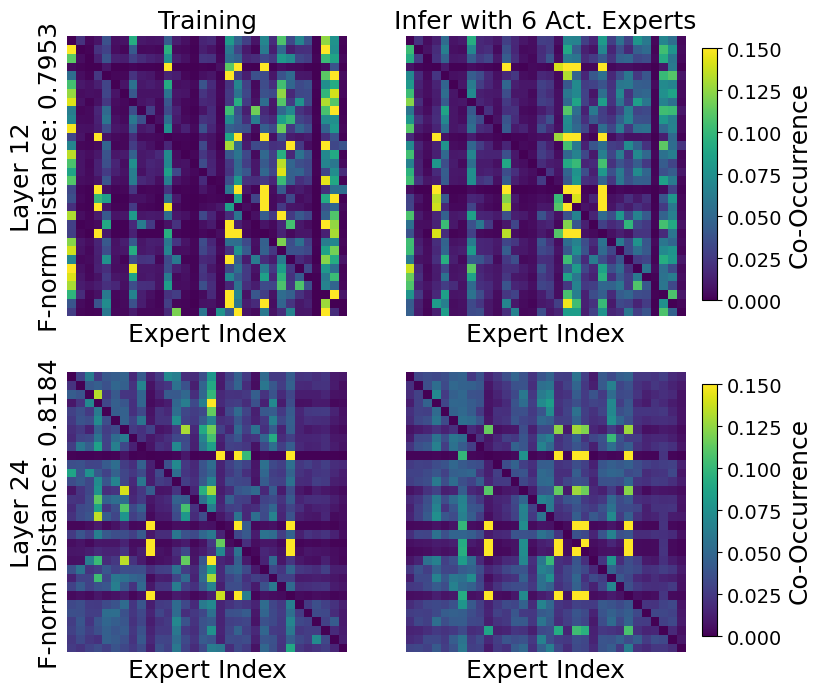}\label{fig:ana3-emoe}}
\caption{Visualization of expert co-occurrence matrices for (a) Top-$k$ and (b) EMoE. We compare the training pattern with inference using $k'=6$, and report the F-norm distance between the corresponding matrices during training and inference. }
\label{fig:ana-3}
\vspace{-0.04cm}
\end{figure*}

\textbf{Effect of Stochastic Co-activation Sampling.} Figure~\ref{fig:ana-3} compares the expert co-occurrence matrices of EMoE and Top-$k$ when extrapolated to $k'=6$. With co-activation sampling, the inference-time co-occurrence matrix maintains high structural similarity to that observed during training. This stability is quantitatively supported by a sharply reduced Frobenius norm distance, which indicate that co-activation sampling effectively learns the expert combination patterns required under higher budgets, thereby ensuring scalability during inference. Appendix~\ref{app:selection-stable} and~\ref{app:diversity} further investigates how $\mathcal{L}_{\text{HR}}$ stabilizes expert selection and evaluates expert diversity.

\begin{wrapfigure}[9]{r}{0.37\textwidth}
\vspace{-0.6cm}
  \centering
  \includegraphics[width=0.36\textwidth]{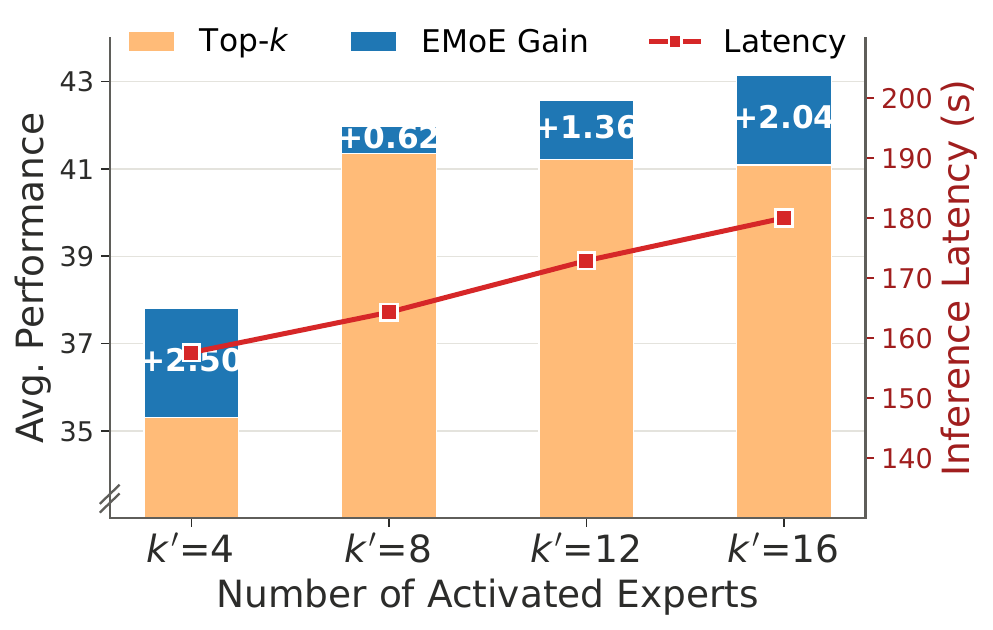}
  \vspace{-0.2cm}
  \caption{Performance and inference latency on OLMoE 
under varying $k'$.}
  \label{fig:overhead}
\end{wrapfigure}
\textbf{Efficiency under Elastic Inference.} We measure end-to-end inference latency on OLMoE (single H200 GPU, 2K input / 2K output tokens).  As shown in 
Figure~\ref{fig:overhead}, scaling from $k'=4$ to $k'=16$ 
incurs moderate latency growth. EMoE converts this additional compute into monotonic performance gains, whereas Top-$k$ peaks at $k'=8$ and declines thereafter, meaning the extra latency is entirely wasted. This confirms that EMoE converts additional compute into 
genuine performance improvements, while standard Top-$k$ 
wastes it due to the collaboration failure identified 
in Section~\ref{sec:pilot}.

\vspace{-0.1cm}
\section{Conclusion}
\vspace{-0.1cm}
Deploying MoE models across diverse computational environments requires inference-time elasticity. In this paper, we identify that existing MoE models suffer from an inference-time scaling wall due to insufficient expert collaboration. To address this issue, we propose Elastic MoE (EMoE), which incorporates stochastic co-activation sampling and hierarchical router loss to break this wall. EMoE enables a single trained MoE model to reliably serve multiple inference budgets, requiring only lightweight post-training on pretrained MoE checkpoints without architectural 
modification. Extensive experiments demonstrate that EMoE extends the effective scaling range to 2--3$\times$ the training-time $k$, achieving monotonic performance gains.



\vspace{-0.1cm}
{\small
\bibliographystyle{unsrtnat}
\bibliography{neurips_2026}
}


\appendix
\vspace{-0.1cm}
\section{Implementation Details}
\label{sec:imp}
\paragraph{Details of Baselines.}
We evaluate our proposed EMoE framework against two primary categories of baselines: standard Top-$k$ routing and dynamic routing methods.

\begin{itemize}
    \item \textbf{Standard Top-$k$ Routing:} This is the most prevalent approach in mainstream MoE models. Crucially, our EMoE framework is trained with the exact same number of activated experts to ensure an identical training overhead.
    \begin{itemize}
        \item For the \textbf{FFN-based} MoE models, we adhere to the official configurations of OLMoE-0924 and DeepSeek-V2-Lite as used during their pre-training.
        \item For the \textbf{LoRA-based} MoE, we adopt the sparse activation pattern commonly used in large-scale models~\citep{deepseekai2025deepseekv3technicalreport}, activating 2 out of 32 total experts (a 6.25\% activation rate).
    \end{itemize}

    \item \textbf{Dynamic Routing Methods:} We compare against two state-of-the-art dynamic routing techniques, Top-p and AdaMoE, which adjust expert activation per token.
    \begin{itemize}
        \item \textbf{Top-$p$ Routing}~\citep{huang-etal-2024-harder} activates the smallest set of experts whose cumulative probability mass exceeds a threshold $p$. To maintain a comparable training budget, we set $p=0.15$ during training. At inference, to match the average expert counts of other methods, we use $p$ values of $\{0.05, 0.16, 0.25, 0.34\}$.
        \item \textbf{AdaMoE}~\citep{DBLP:conf/emnlp/ZengMG0D24} introduces null experts that can be routed to, effectively allowing the model to skip computation for certain tokens. Following the original implementation, we set the number of null experts to be twice that of the standard experts. For a fair inference-time comparison, we vary its target active expert count $k'$ to $\{3, 6, 14, 22\}$ to align its average number of activated non-null experts with the computational budgets of Top-$k$ and EMoE.
    \end{itemize}
\end{itemize}
\paragraph{Details of Training Hyperparameters.}  
In our main experiments, the learning rate is set to $2 \times 10^{-4}$ for all methods under the LoRA-based settings, and $2 \times 10^{-5}$ under the FFN-based settings. We use a batch size of 128 in all cases. All models are fine-tuned for 4 epochs on the dataset with a sequence length of 2048. For the hierarchical loss coefficient $\lambda$, we use a value of $5 \times 10^{-4}$ in LoRA-based scenarios and $1 \times 10^{-8}$ in FFN-based scenarios. The gap arises because pretrained FFN-based routers already carry learned routing patterns, so a small $\lambda$ suffices 
to further refine the ranking, whereas LoRA routers are trained from scratch with near-uniform initialization and require a larger $\lambda$ to establish decisive rankings. Within each category, the same $\lambda$ is used across all models without per-model tuning. Experiments are performed on 8 Nvidia H100 GPUs, each equipped with 80GB of memory. Every experiment is repeated three times, and we report the mean and standard deviation of the results.

\paragraph{Details of Evaluation.}
We conduct a comprehensive evaluation utilizing the OpenCompass package~\citep{2023opencompass} to assess model performance across a diverse suite of downstream benchmarks. We report zero-shot accuracy on the commonsense and multitask reasoning tasks ARC-e, ARC-c~\citep{DBLP:journals/corr/abs-1803-05457}, MMLU~\citep{DBLP:conf/iclr/HendrycksBBZMSS21}, and WinoGrande~\citep{DBLP:conf/aaai/SakaguchiBBC20}. For reasoning capabilities, we measure 8-shot accuracy on the mathematical reasoning benchmark GSM8K~\citep{DBLP:journals/corr/abs-2110-14168} and 3-shot accuracy on HellaSwag~\citep{DBLP:conf/acl/ZellersHBFC19}. Coding is evaluated via the pass@1 metric on HumanEval~\citep{DBLP:journals/corr/abs-2107-03374}. To complete the assessment, our evaluation also includes two prominent open-domain question-answering benchmarks, Natural Questions~\citep{DBLP:journals/tacl/KwiatkowskiPRCP19} and TriviaQA~\citep{DBLP:conf/acl/JoshiCWZ17}.
\vspace{-0.1cm}
\section{Extended Experiments}
\label{sec:app_exp}
\subsection{Comparison to Training with Larger \texorpdfstring{$k_{\text{train}}$}{k\_train}}
\label{app:large-k}
We compare EMoE trained with $k_{\text{train}}=2$ against a standard Top-$k$ MoE 
trained with $k_{\text{train}}=6$ in Table~\ref{tab:large-k}. The large-$k_{\text{train}}$ model achieves only a marginal advantage over EMoE without $\mathcal{L}_\text{HR}$ at its native inference 
budget ($k'=6$), but exhibits two clear drawbacks: (1) Lack of elasticity. Performance degrades sharply when reducing the inference budget 
(e.g., $k'=2 < k_{\text{train}}$), showing that standard MoE training 
strongly couples the router to the training-time budget. (2) Excessive training cost. Beyond the standard $k$ 
configuration (e.g., from 2 to 6) requires roughly 3$\times$ FLOPs and more activation memory in MoE layers, making such large-$k$ training impractical at scale. 

In contrast, EMoE maintains strong performance under both lower inference budgets 
($k' < k_{\text{train}}$) and higher inference budgets ($k' > k_{\text{train}}$),
while retaining the training cost of standard Top-$k$ models.
Thus, EMoE provides inference-time elasticity that large-$k_{\text{train}}$
training is unable to offer and eliminates the need to deploy 
multiple MoE variants for different computational budgets.
\begin{table*}[t]
\centering
\caption{Comparison between EMoE and Top-$k$ trained with large $k_{\text{train}}$.}
\begin{tabular}{lccccc}
\toprule
Model & $k'=1$ & $k'=2$ & $k'=4$ & $k'=6$ & Mem / GPU \\
\midrule
Top-$k$ ($k_{\text{train}}=6$) & \textcolor{red!50!black}{42.87} & \textcolor{red!50!black}{46.26} & 48.30 & 49.14 & \textcolor{red!50!black}{76.0GB} \\
EMoE (w/o $\mathcal{L}_{\text{HR}}$, $k_{\text{train}}=2$) 
& 45.19 & 47.79 & 48.81 & 49.08 & 43.4GB \\
EMoE (with $\mathcal{L}_{\text{HR}}$, $k_{\text{train}}=2$) 
& 45.68 & 48.22 & 49.00 & 49.50 & 43.4GB \\
\bottomrule
\end{tabular}
\label{tab:large-k}
\end{table*}

\subsection{Effect of Larger Training Budgets}
\label{app:ktrain-scaling}
\begin{wrapfigure}[13]{r}{0.35\textwidth}
\vspace{-0.45cm}
  \centering
  \includegraphics[width=0.35\textwidth]{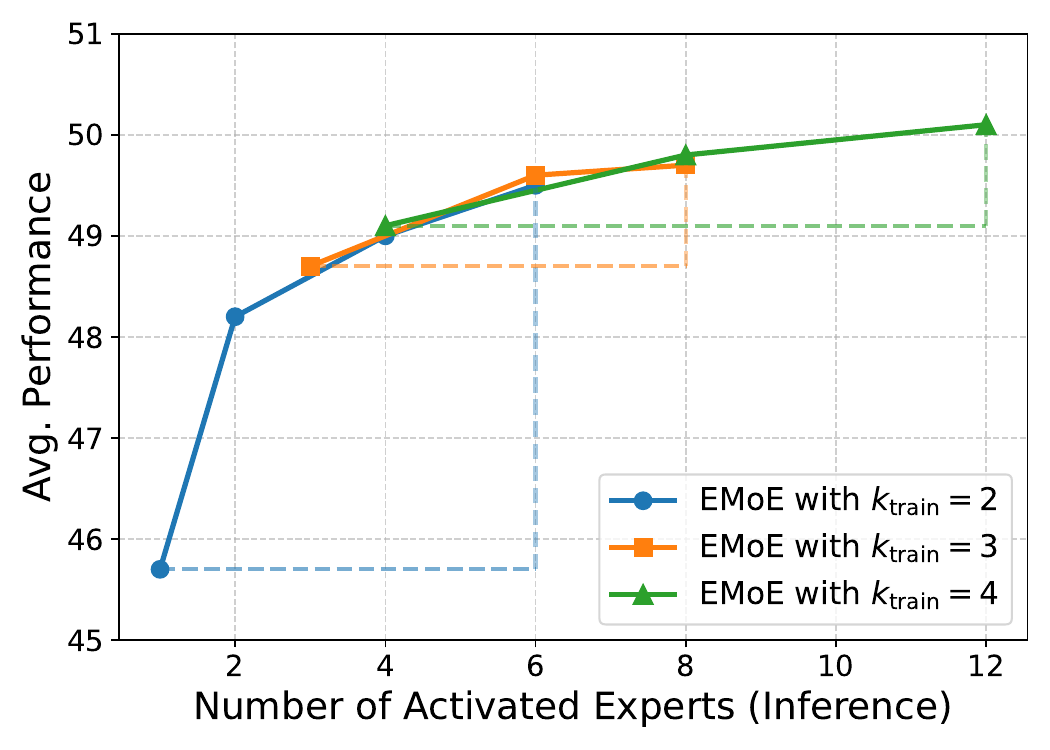}
  \caption{Average performance of EMoE for different training-time budgets $k_{\text{train}} \in \{2,3,4\}$.}
  \label{fig:ktrain-scaling}
\end{wrapfigure}
To study whether EMoE continues to benefit from additional training compute, 
we further vary the training-time budget $k_{\text{train}}$ and 
compare models trained with $k_{\text{train}} \in \{2,3,4\}$. 
For each setting, we evaluate the resulting EMoE model under multiple inference budgets $k'$ 
and report the average performance across the same evaluation suite as in the main experiments. 
As shown in Figure~\ref{fig:ktrain-scaling}, we observe two trends. 
First, for each $k_{\text{train}}$, EMoE maintains an elastic regime in which performance 
improves smoothly as $k'$ increases beyond $k_{\text{train}}$. 
Second, increasing $k_{\text{train}}$ consistently lifts the entire curve, especially at larger 
inference budgets. 
This shows that EMoE continues to benefit from additional training compute, while preserving its elasticity across $k'$.
\subsection{Effect of Larger Datasets}
\label{app:larger-data}
We extend our experiments to 
larger 100K and 200K training datasets. The results are reported in 
Table~\ref{tab:100k-200k}. Across both data scales, EMoE consistently preserves its elastic window and maintains strong extrapolation capability beyond the training-time $k_{\text{train}}$. Importantly, enlarging the dataset does not diminish the benefits of EMoE: the method continues to outperform standard Top-$k$ training at both lower 
($k' < k_{\text{train}}$) and higher ($k' > k_{\text{train}}$) inference budgets. These results demonstrate that EMoE is a 
lightweight, plug-and-play adaptation for pretrained MoE models: 
its effectiveness is independent of dataset scale and can be 
achieved efficiently with a lightweight 50K instruction set.
\begin{table*}[t]
\caption{Comparison between EMoE and standard Top-$k$ across inference-time budgets $k'$
on larger instruction-tuning datasets (100K and 200K samples).
For each dataset size, both methods are trained under the same $k_{\text{train}} = 2$.}
\small
\centering
\setlength{\tabcolsep}{2.5pt}
\begin{tabular}{@{}lccccccccc>{\columncolor{gray!20}}cc@{}@{}}
\toprule
& \textbf{ARC-c} & \textbf{ARC-e} & \textbf{GSM8K} & \textbf{HellaS.} & \textbf{HumanE.} & \textbf{NQ} & \textbf{Tri.QA} & \textbf{Wino.} & \textbf{MMLU} & \textbf{AVG} \\ \midrule
\midrule
\multicolumn{11}{l}{\textit{100K instruction data}} \\
Top-$k$ ($k' = 1$) & 53.90 & 71.25 & 34.19 & 58.47 & 20.12 & 21.99 & 52.70 & 55.41 & 46.05 & 46.01 \\
Top-$k$ ($k' = 2$) & 54.24 & 75.66 & 42.00 & 56.67 & 25.61 & 24.82 & 57.49 & 54.38 & 46.66 & 48.61 \\
Top-$k$ ($k' = 4$) & 55.93 & 77.78 & 43.75 & 55.53 & 26.83 & 27.37 & 59.14 & 55.49 & 45.38 & 49.69 \\
Top-$k$ ($k' = 6$) & 56.27 & 77.25 & 40.33 & 53.51 & 21.34 & 28.42 & 59.88 & 54.62 & 46.45 & 48.67 \\
\midrule
EMoE ($k' = 1$) & 57.29 & 76.54 & 37.45 & 57.01 & 14.02 & 24.07 & 54.76 & 53.51 & 46.57 & \textbf{46.80}\\
EMoE ($k' = 2$) & 60.00 & 79.72 & 40.64 & 59.47 & 17.07 & 26.40 & 58.45 & 55.33 & 47.86 & \textbf{49.44} \\
EMoE ($k' = 4$) & 62.71 & 80.95 & 41.09 & 61.04 & 20.12 & 27.09 & 59.57 & 55.56 & 48.86 & \textbf{50.78} \\
EMoE ($k' = 6$) & 63.05 & 80.95 & 42.30 & 61.13 & 21.34 & 27.42 & 60.09 & 56.20 & 48.87 & \textbf{51.26} \\
\midrule
\midrule
\multicolumn{11}{l}{\textit{200K instruction data}} \\
Top-$k$ ($k' = 1$) & 58.31 & 76.54 & 40.03 & 61.26 & 27.44 & 23.13 & 53.92 & 52.41 & 48.35 & 49.04 \\
Top-$k$ ($k' = 2$) & 60.34 & 79.37 & 44.66 & 63.92 & 29.88 & 25.60 & 58.09 & 53.20 & 50.09 & 51.68 \\
Top-$k$ ($k' = 4$) & 61.02 & 81.13 & 44.96 & 63.46 & 28.66 & 27.40 & 59.91 & 54.14 & 50.66 & 52.37 \\
Top-$k$ ($k' = 6$) & 59.32 & 81.31 & 43.52 & 62.11 & 25.61 & 27.34 & 60.29 & 53.67 & 50.27 & 51.49 \\
\midrule
EMoE ($k' = 1$) & 62.71 & 75.13 & 42.30 & 60.98 & 21.95 & 25.04 & 56.08 & 55.33 & 47.51 & \textbf{49.67} \\
EMoE ($k' = 2$) & 65.08 & 78.84 & 44.28 & 62.31 & 25.61 & 25.57 & 58.67 & 58.56 & 49.19 & \textbf{52.01} \\
EMoE ($k' = 4$) & 64.75 & 81.31 & 46.47 & 63.29 & 22.56 & 27.70 & 60.20 & 58.17 & 50.21 & \textbf{52.74} \\
EMoE ($k' = 6$) & 66.44 & 78.13 & 47.38 & 63.60 & 25.61 & 27.98 & 60.36 & 57.14 & 51.99 & \textbf{53.18} \\ \bottomrule
\end{tabular}
\label{tab:100k-200k}
\end{table*}

\section{Extended Analysis}
\subsection{Relationship between $\mathcal{L}_\text{HR}$ 
and Load Balancing Loss}
\label{app:lbl}
\begin{table}[t]
\small
\centering
\caption{Ablation on load balancing loss (LoRAMoE).}
\label{tab:lbl-ablation}
\begin{tabular}{lcccc}
\toprule
 & $k'$=1 & $k'$=2 & $k'$=4 & $k'$=6 \\
\midrule
Top-$k$ (w/o $\mathcal{L}_b$) & 44.39 & 47.46 & 47.97 & 47.31 \\
Top-$k$ (w/ $\mathcal{L}_b$) & 45.48 & 47.88 & 48.05 & 47.57 \\
Top-$k$ (w/ $\mathcal{L}_b$ + $\mathcal{L}_\text{HR}$) & 45.83 & 48.03 & 48.68 & 48.15 \\
\bottomrule
\end{tabular}
\end{table}
A natural question is whether $\mathcal{L}_\text{HR}$ conflicts with the load balancing loss $\mathcal{L}_b$. 
To investigate this, we ablate $\mathcal{L}_b$ and compare 
performance across inference budgets. As shown in Table~\ref{tab:lbl-ablation}, removing $\mathcal{L}_b$ degrades performance at every 
$k'$ and the scaling wall persists (performance drops 
from $k'=4$ to $k'=6$). Adding $\mathcal{L}_\text{HR}$ 
on top of $\mathcal{L}_b$ yields consistent improvements 
across all budgets, including at $k'=6$ where it gains 
+0.58 over standard Top-$k$. This confirms that 
$\mathcal{L}_\text{HR}$ provides a qualitatively different 
benefit from $\mathcal{L}_b$: the former sharpens 
within-token expert ranking, while the latter balances 
cross-token expert usage. Their effects are complementary 
rather than conflicting.
\subsection{Effect of $k_\text{ideal}$ on DeepSeek-V2-Lite}
\label{app:ideal-dsv2}
\begin{wrapfigure}[13]{r}{0.40\textwidth}
\vspace{-0.7cm}
  \centering
  \includegraphics[width=0.40\textwidth]{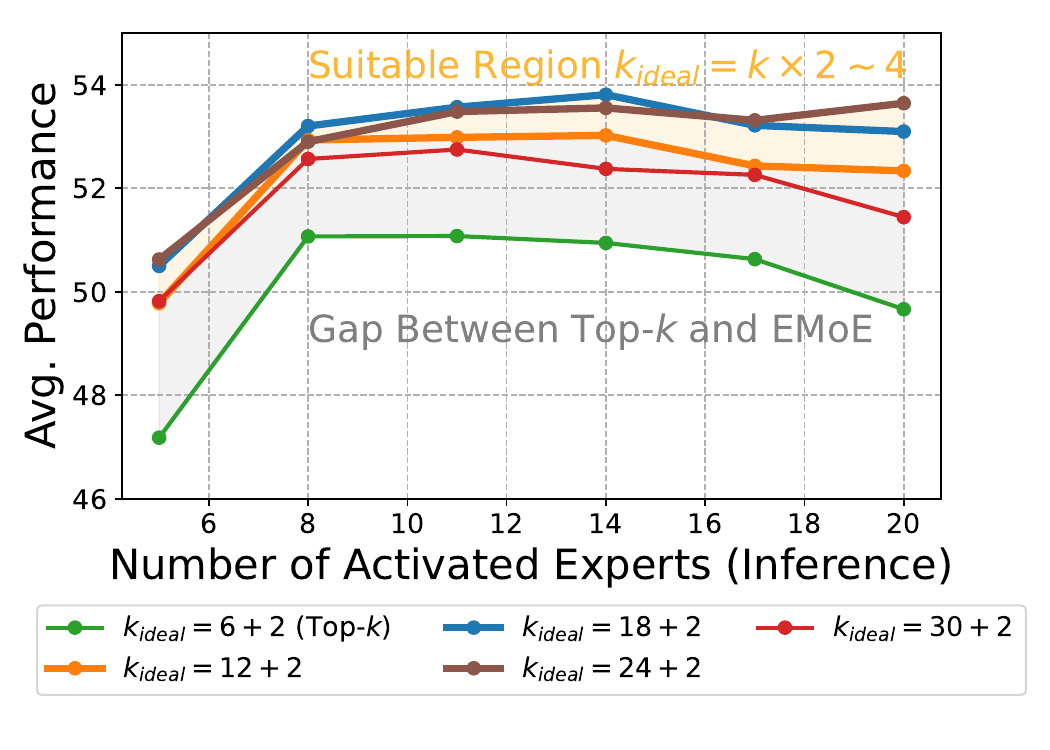}
  \vspace{-0.8cm}
  \caption{Analysis of the effect of the hyperparameter $k_\text{ideal}$ on DeepSeek-V2.}
  \label{fig:app-ana-2}
\end{wrapfigure}
We conduct an analysis of the hyperparameter $k_{\text{ideal}}$ on the DeepSeek-V2-Lite model to further validate the robustness of its configuration. The results in Figure~\ref{fig:app-ana-2} clearly demonstrate that the choice of $k_{\text{ideal}}$ offers considerable elasticity and tolerance. Consistent with the conclusions drawn from Figure~\ref{fig:ana-2}, setting $k_{\text{ideal}}$ to only twice the number of training experts ($k_{\text{train}} = 12+2$) leads to significant improvements in the performance of EMoE, exceeding the standard Top-$k$ in both peak and low-budget scenarios. Furthermore, when $k_{\text{ideal}}$ is set within 2 to 4 times the number of training experts, the model achieves optimal performance and scalability, reaching the highest average scores within this range.

\subsection{Effect of $\mathcal{L}_\text{HR}$ on Expert Selection Stability}
\label{app:selection-stable}
\begin{figure*}[t]
\centering
\subfloat[Math]{
    \includegraphics[width=0.45\linewidth]{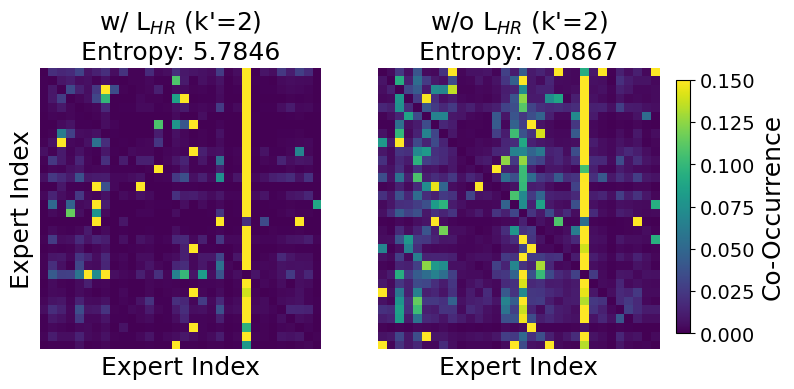}
    \label{fig:lhr-math}
}
\hspace{5mm}
\subfloat[Commonsense]{
    \includegraphics[width=0.45\linewidth]{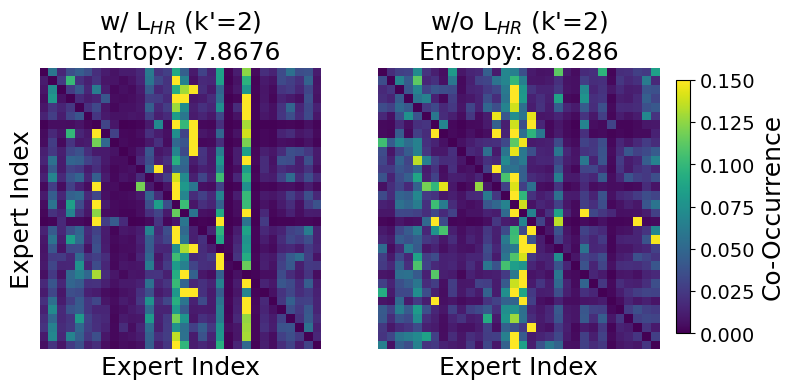}
    \label{fig:lhr-common}
}
\caption{
Expert co-occurrence visualization under a low inference budget ($k' = 2$) across two domains.  
Across domains, models \textbf{with $\mathcal{L}_{\text{HR}}$} exhibit sparse, concentrated hotspots, 
while models \textbf{without $\mathcal{L}_{\text{HR}}$} show diffuse patterns with higher uncertainty.}
\label{fig:lhr-cooccur}
\end{figure*}
To verify the effect of $\mathcal{L}_{\text{HR}}$ on selecting more favorable expert combinations, 
we further visualize expert co-occurrence matrices under a low inference budget ($k'=2$) 
on two domains: math and commonsense, using GSM8K~\citep{DBLP:journals/corr/abs-2110-14168} and HellaSwag~\citep{DBLP:conf/acl/ZellersHBFC19} respectively. 
As shown in Figure~\ref{fig:lhr-cooccur}, models \emph{without} $\mathcal{L}_{\text{HR}}$ exhibit diffuse, 
weakly structured co-activation patterns, indicating unstable selection when only a few experts 
can be used. In contrast, models \emph{with} $\mathcal{L}_{\text{HR}}$ display sparse, concentrated hotspots, 
evidence of more decisive and consistent expert selection. 
This sharpening effect is also reflected quantitatively: entropy decreases from 7.09 to 5.78 (math) 
and from 8.63 to 7.87 (commonsense). 
These results show that $\mathcal{L}_{\text{HR}}$ significantly stabilizes routing and 
is important for elasticity.

\subsection{Comparison with Expert Dropout}
\label{app:dropout}
\begin{table}[t]
\centering
\small
\caption{Comparison with expert-level dropout ($p$=0.2).}
\label{tab:dropout}
\begin{tabular}{lcccc}
\toprule
Method & $k'=1$ & $k'=2$ & $k'=4$ & $k'=6$ \\
\midrule
Top-$k$ & 45.48 & 47.88 & 48.05 & 47.57 \\
Top-$k$ + Dropout & \textbf{46.07} & 47.63 & 47.71 & 47.11 \\
EMoE (Ours) & 45.68 & \textbf{48.22} & \textbf{49.00} & \textbf{49.50} \\
\bottomrule
\end{tabular}
\end{table}
Our stochastic co-activation sampling may appear related to 
expert-level dropout, but differs in motivation and mechanism. 
Dropout is ``subtractive'': randomly dropping experts from 
top-$k$ for regularization within a fixed budget. Our approach 
is ``additive'': sampling from a larger candidate pool 
($k_{\text{ideal}} > k_{\text{train}}$) to expose the model 
to combinations that emerge under higher inference budgets. 
Dropout does not address the collaboration gap between training 
and inference, whereas our method explicitly prepares the model 
for variable budgets.

To empirically validate the distinction between our method 
and dropout-based regularization, we compare EMoE with a 
Top-$k$ baseline augmented with expert dropout during training. 
As shown in Table~\ref{tab:dropout}, while dropout provides 
marginal improvement at $k'=1$ due to its regularization effect, 
it fails to address the scaling wall: performance still degrades 
at $k'=6$. In contrast, EMoE achieves monotonically increasing 
performance across inference budgets. This confirms that 
dropout's ``subtractive'' mechanism cannot substitute for our 
``additive'' co-activation sampling, which explicitly trains 
expert combinations that emerge at higher inference budgets.

\begin{table}[t]
\centering
\small
\caption{Mutual information between expert usage and task domain.}
\begin{tabular}{lcc}
\toprule
Model Setup & MI & $\Delta$ \\
\midrule
Baseline (Standard Top-$k$) & 0.0473 & -- \\
+ Co-activation Sampling (w/o $\mathcal{L}_{\text{HR}}$) & 0.0603 & +27.5\% \\
+ Full EMoE (with $\mathcal{L}_{\text{HR}}$) & \textbf{0.0630} & +4.5\% \\
\bottomrule
\end{tabular}
\label{tab:mi}
\end{table}
\subsection{Analysis of Expert Diversity}
\label{app:diversity}
To examine how the two components in our method influence expert specialization,
we measure the mutual information (MI) between expert selection and task domains
(math: GSM8K~\citep{DBLP:journals/corr/abs-2110-14168}, commonsense: HellaSwag~\citep{DBLP:conf/acl/ZellersHBFC19}, code: HumanEval~\citep{DBLP:journals/corr/abs-2107-03374}).
Let $P(e)$ denote the marginal expert usage and $P(e \mid d)$ the domain-conditional usage.
We compute:
\begin{equation}
\mathrm{MI}(E;D)
= \sum_{e,d} P(e,d)\,\log\frac{P(e,d)}{P(e)P(d)},
\end{equation}
where $P(e,d)=P(e \mid d)P(d)$. As shown in Table~\ref{tab:mi}, co-activation sampling is the primary factor that enhances specialization by exposing 
experts to diverse collaborative configurations during training. 
A higher degree of specialization indicates that experts consistently assume distinct functional roles across domains, which is an important symptom of successful collaborative organization rather than redundant or interchangeable behaviors. The hierarchical router loss further improves this structure by producing more decisive expert rankings, achieving the highest MI.
These findings show that $\mathcal{L}_{\text{HR}}$ works synergistically with co-activation sampling 
and plays a central role in EMoE’s elasticity.

\subsection{Training Dynamics}
\begin{figure*}[t]
\centering
\subfloat[Epoch 1 checkpoint]{	\includegraphics[width=0.24\linewidth]{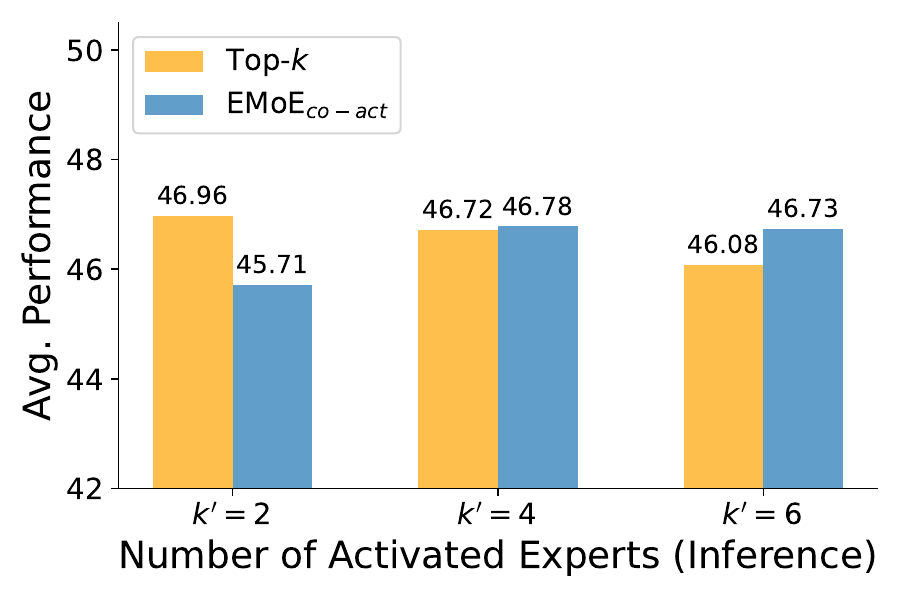}} 
\subfloat[Epoch 2 checkpoint]{ \includegraphics[width=0.24\linewidth]{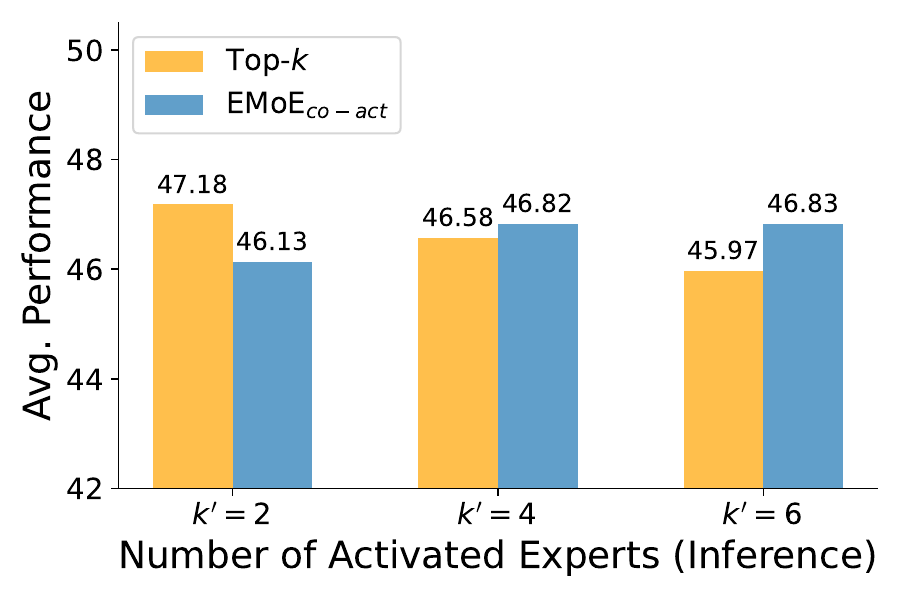}} 
\hfill
\subfloat[Epoch 3 checkpoint]{ \includegraphics[width=0.24\linewidth]{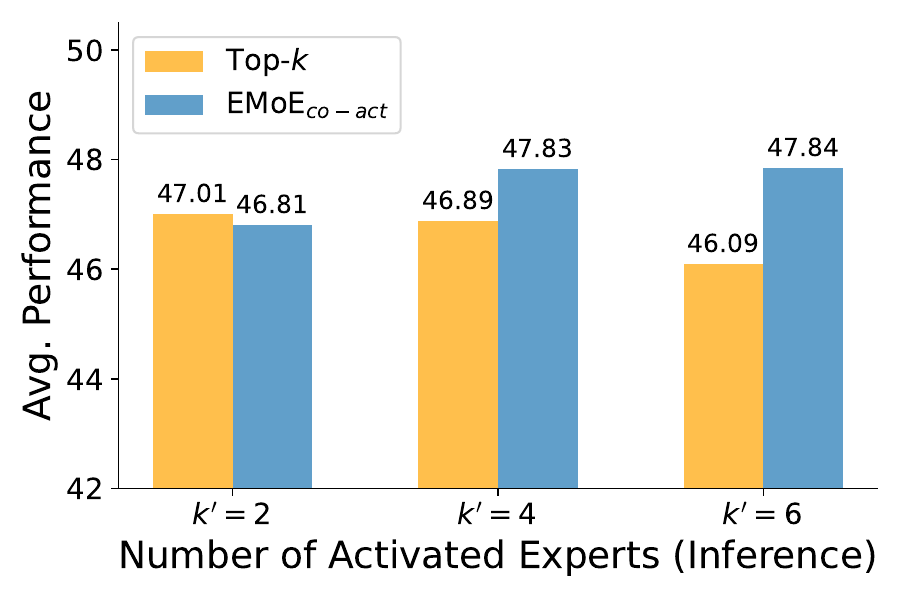}} 
\subfloat[Epoch 4 checkpoint]{ \includegraphics[width=0.24\linewidth]{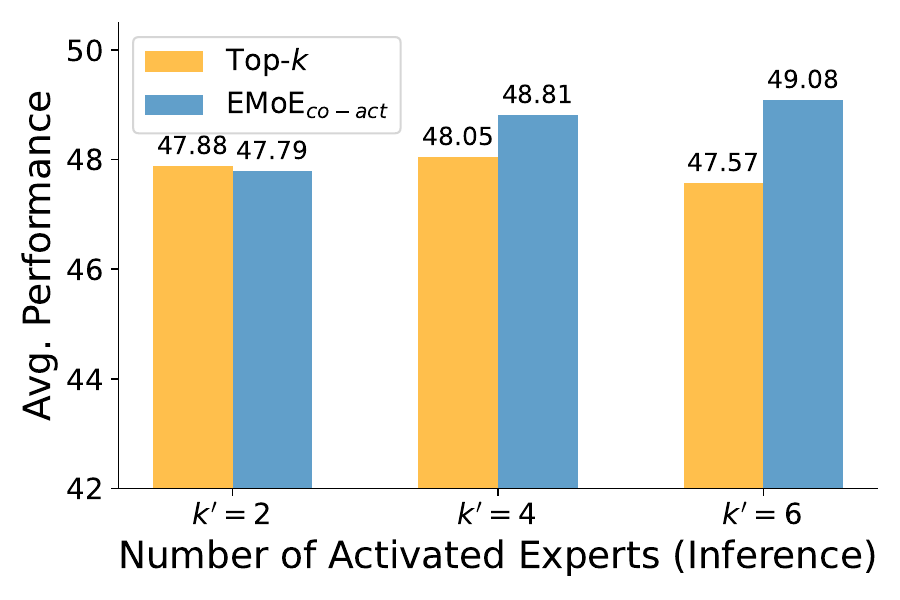}}
\caption{Performance evolution of EMoE$_\text{co-act}$ (i.e., only using co-activation sampling) versus the Top-$k$ baseline at different training checkpoints. The subplots (a) through (d) show performance snapshots at the end of epochs 1, 2, 3, and 4, respectively.}
\label{fig:training_dynamics}
\end{figure*}
To gain deeper insight into the learning process of EMoE and investigate the impact of co-activation sampling on performance, we evaluate the model’s performance at the end of each training epoch and compare it to the Top-$k$ baseline. Figure \ref{fig:training_dynamics} illustrates this dynamic process. The experiments reveal a key trade-off. In the early stage of training (epoch 1), when only a small number of experts are activated during inference ($k'=2$), EMoE with co-activation sampling only (i.e., EMoE$_\text{co-act}$) underperforms the Top-k baseline. We believe this is because the random sampling mechanism forces the model to explore a broader range of expert combinations, thus dispersing learning resources away from optimizing the most frequent Top-2 combinations. This leads to slightly slower convergence in this specific setting. However, even at this stage, our method already begins to outperform the Top-$k$ method in broader activation regimes ($k'=4$ and $k'=6$), indicating that the model has started to learn how to leverage more experts in collaboration.
\begin{table}[t]
\caption{Training overhead comparison between EMoE and the Top-$k$ baseline under the LoRA-based settings used in our main experiments. Both methods are trained with $k_{\text{train}}=2$ on the same hardware.}
\centering
\small
\begin{tabular}{@{}lccc@{}}
\toprule
Method & $k_\text{train}$ & Training Time & Memory Usage Per GPU \\
\midrule
EMoE & 2 & 10.92h & 43.4GB \\
Top-$k$ & 2 & 10.92h & 43.4GB \\
\bottomrule
\end{tabular}
\label{tab:training_budget}
\end{table}

As training progresses, this early trade-off is perfectly resolved. From epoch 2 onwards, EMoE$_\text{co-act}$ consistently matches or surpasses the baseline across all inference configurations. By the end of training (epoch 4), both models converge to their optimal performance, but with markedly different results. EMoE$_\text{co-act}$ achieves optimal performance under all inference budgets: not only does it match the Top-$k$ performance in the standard $k'=2$ setting, but more importantly, it successfully extends this optimization to all activation regimes, exhibiting strong and monotonic performance scalability. In contrast, although the Top-$k$ baseline also converges to its optimal performance at $k'=2$, its performance curve demonstrates that it fails to learn how to utilize additional experts.

\subsection{Training Efficiency}
\label{app:training-eff}
An essential consideration for the EMoE framework is its computational efficiency during the training process. To quantify its overhead, we compare it with the standard Top-$k$ baseline method. As shown in Table~\ref{tab:training_budget}, EMoE achieves the same training overhead as the Top-$k$ baseline. This is because the core components of EMoE are introduced in the non-dense computation part of the computation graph. Specifically, both the sampling of experts from a larger candidate pool $\mathcal{S}_{k_{\text{ideal}}}(x)$ and the calculation of the hierarchical loss incur negligible computational cost compared to the dense matrix operations in Transformer models.  Overall, EMoE successfully unlocks elastic scalability during inference with no extra training overhead, demonstrating its practical value as a lightweight and efficient training framework.

\section{Algorithm of EMoE}
Here, we present the complete algorithm of the proposed EMoE training framework in Algorithm~\ref{alg:emoe}.
\begin{algorithm}[H]
\caption{Elastic Mixture-of-Experts (EMoE) Training Framework}
\label{alg:emoe}
\begin{algorithmic}[1]
\State  \textbf{Require:} Input $x$, Router $G$, Experts $\{E_i(\cdot;\theta_i)\}_{i=1}^N$
\State  \textbf{Hyperparameters:} $k_{\text{ideal}}, \lambda$
\item[]

\State  \textbf{Step 1: Get router logits}
\State  $h(x) \gets G(x)$
\Comment{Raw logits for all experts, $h(x) \in \mathbb{R}^N$}
\item[]

\State  \textbf{Step 2: Stochastic co-activation sampling}
\State  \textbf{Step 2a: Determine candidate pool size}
\State  $\tilde{k}_{\text{ideal}} \sim \UniformInt(k_{\text{train}}, k_{\text{ideal}})$
\State  $\mathcal{S}_{\tilde{k}_{\text{ideal}}}(x) \gets \TopKIndices(h(x), \tilde{k}_{\text{ideal}})$
\Comment{Select candidate experts based on top logits}

\State  \textbf{Step 2b: Sample experts for forward pass}
\State  $\mathcal{S}_{\text{co-act}}(x) \sim \UniformSample(\mathcal{S}_{\tilde{k}_{\text{ideal}}}(x), k_{\text{train}})$
\Comment{Final subset used for training}
\item[]

\State  \textbf{Step 3: Compute MoE output}
\State  $y_{\text{co-act}}(x) \gets \sum_{i \in \mathcal{S}_{\text{co-act}}(x)} 
\frac{\exp(h_i(x))}{\sum_{j \in \mathcal{S}_{\text{co-act}}(x)} \exp(h_j(x))} \cdot E_i(x; \theta_i)$
\item[]

\State  \textbf{Step 4: Compute total loss}
\State  $\mathcal{L}_{\text{ce}} \gets \text{CrossEntropyLoss}(y_{\text{co-act}}(x), \text{target})$
\State  $\mathcal{L}_{\text{b}} \gets \text{LoadBalancingLoss}(h(x))$
\State $q(x) \gets \mathrm{softmax}(h(x))$
\State $\mathcal{L}_{\text{HR}}(x) \gets
-\sum_{i=1}^{N} q_i(x)\log\frac{q_i(x)}{1/N}$
\Comment{Hierarchical router loss (encourage decisive ranking)}
\State  $\mathcal{L}_{\text{total}} \gets \mathcal{L}_{\text{ce}} + \mathcal{L}_{\text{b}} + \lambda \cdot \mathcal{L}_{\text{HR}}$
\State  \textbf{return} $\mathcal{L}_{\text{total}}$
\end{algorithmic}
\end{algorithm}

\section{Limitations}
Although EMoE demonstrates strong effectiveness across various model sizes and architectures, we acknowledge two limitations. First, the elastic range is bounded. Our analysis of $k_{\text{ideal}}$ in Figure~\ref{fig:ana-2} shows that setting $k_{\text{ideal}}$ within 2 to 4 times $k_{\text{train}}$ achieves optimal performance, but excessively large values lead to diminishing returns. In the extreme case where $k_{\text{ideal}} = N$ with $k_{\text{train}} = 2$, this reduces to randomly selecting two experts, rendering the router ineffective. We will explore approaches to further extend this effective range in future work. Second, our validation on ultra-large-scale models is limited. Due to computational constraints, our experiments focus on models ranging from 7B to 21B parameters. While EMoE demonstrates consistent scalability across these sizes, we have not yet evaluated it on models exceeding 100B parameters, which represents an important direction for future research.

\section{Broader Impact}
EMoE enables a single trained MoE model to serve across heterogeneous hardware and fluctuating workloads, reducing the need to train and maintain multiple model variants. This can lower the computational and energy costs of large-scale deployments, making high-performance models more accessible to researchers and organizations with limited resources. We do not foresee negative societal impacts specific to this work beyond those generally associated with improving the efficiency of large language models.


\newpage
\section*{NeurIPS Paper Checklist}

\begin{enumerate}

\item {\bf Claims}
    \item[] Question: Do the main claims made in the abstract and introduction accurately reflect the paper's contributions and scope?
    \item[] Answer: \answerYes{} 
    \item[] Justification: The main claims made in the abstract and introduction (Section 1) accurately reflect the paper’s contributions and scope.
    \item[] Guidelines:
    \begin{itemize}
        \item The answer \answerNA{} means that the abstract and introduction do not include the claims made in the paper.
        \item The abstract and/or introduction should clearly state the claims made, including the contributions made in the paper and important assumptions and limitations. A \answerNo{} or \answerNA{} answer to this question will not be perceived well by the reviewers. 
        \item The claims made should match theoretical and experimental results, and reflect how much the results can be expected to generalize to other settings. 
        \item It is fine to include aspirational goals as motivation as long as it is clear that these goals are not attained by the paper. 
    \end{itemize}

\item {\bf Limitations}
    \item[] Question: Does the paper discuss the limitations of the work performed by the authors?
    \item[] Answer: \answerYes{} 
    \item[] Justification: The paper discusses the limitations of the approach in the Appendix E.
    \item[] Guidelines:
    \begin{itemize}
        \item The answer \answerNA{} means that the paper has no limitation while the answer \answerNo{} means that the paper has limitations, but those are not discussed in the paper. 
        \item The authors are encouraged to create a separate ``Limitations'' section in their paper.
        \item The paper should point out any strong assumptions and how robust the results are to violations of these assumptions (e.g., independence assumptions, noiseless settings, model well-specification, asymptotic approximations only holding locally). The authors should reflect on how these assumptions might be violated in practice and what the implications would be.
        \item The authors should reflect on the scope of the claims made, e.g., if the approach was only tested on a few datasets or with a few runs. In general, empirical results often depend on implicit assumptions, which should be articulated.
        \item The authors should reflect on the factors that influence the performance of the approach. For example, a facial recognition algorithm may perform poorly when image resolution is low or images are taken in low lighting. Or a speech-to-text system might not be used reliably to provide closed captions for online lectures because it fails to handle technical jargon.
        \item The authors should discuss the computational efficiency of the proposed algorithms and how they scale with dataset size.
        \item If applicable, the authors should discuss possible limitations of their approach to address problems of privacy and fairness.
        \item While the authors might fear that complete honesty about limitations might be used by reviewers as grounds for rejection, a worse outcome might be that reviewers discover limitations that aren't acknowledged in the paper. The authors should use their best judgment and recognize that individual actions in favor of transparency play an important role in developing norms that preserve the integrity of the community. Reviewers will be specifically instructed to not penalize honesty concerning limitations.
    \end{itemize}

\item {\bf Theory assumptions and proofs}
    \item[] Question: For each theoretical result, does the paper provide the full set of assumptions and a complete (and correct) proof?
    \item[] Answer: \answerNA{} 
    \item[] Justification: This paper is mainly based on observation, making conjectures and methods and proving the effects through experiments.
    \item[] Guidelines:
    \begin{itemize}
        \item The answer \answerNA{} means that the paper does not include theoretical results. 
        \item All the theorems, formulas, and proofs in the paper should be numbered and cross-referenced.
        \item All assumptions should be clearly stated or referenced in the statement of any theorems.
        \item The proofs can either appear in the main paper or the supplemental material, but if they appear in the supplemental material, the authors are encouraged to provide a short proof sketch to provide intuition. 
        \item Inversely, any informal proof provided in the core of the paper should be complemented by formal proofs provided in appendix or supplemental material.
        \item Theorems and Lemmas that the proof relies upon should be properly referenced. 
    \end{itemize}

    \item {\bf Experimental result reproducibility}
    \item[] Question: Does the paper fully disclose all the information needed to reproduce the main experimental results of the paper to the extent that it affects the main claims and/or conclusions of the paper (regardless of whether the code and data are provided or not)?
    \item[] Answer: \answerYes{} 
    \item[] Justification: The paper provides a detailed description of the experimental data setup and hyperparameter settings in Section 5 and the Appendix A.
    \item[] Guidelines:
    \begin{itemize}
        \item The answer \answerNA{} means that the paper does not include experiments.
        \item If the paper includes experiments, a \answerNo{} answer to this question will not be perceived well by the reviewers: Making the paper reproducible is important, regardless of whether the code and data are provided or not.
        \item If the contribution is a dataset and\slash or model, the authors should describe the steps taken to make their results reproducible or verifiable. 
        \item Depending on the contribution, reproducibility can be accomplished in various ways. For example, if the contribution is a novel architecture, describing the architecture fully might suffice, or if the contribution is a specific model and empirical evaluation, it may be necessary to either make it possible for others to replicate the model with the same dataset, or provide access to the model. In general. releasing code and data is often one good way to accomplish this, but reproducibility can also be provided via detailed instructions for how to replicate the results, access to a hosted model (e.g., in the case of a large language model), releasing of a model checkpoint, or other means that are appropriate to the research performed.
        \item While NeurIPS does not require releasing code, the conference does require all submissions to provide some reasonable avenue for reproducibility, which may depend on the nature of the contribution. For example
        \begin{enumerate}
            \item If the contribution is primarily a new algorithm, the paper should make it clear how to reproduce that algorithm.
            \item If the contribution is primarily a new model architecture, the paper should describe the architecture clearly and fully.
            \item If the contribution is a new model (e.g., a large language model), then there should either be a way to access this model for reproducing the results or a way to reproduce the model (e.g., with an open-source dataset or instructions for how to construct the dataset).
            \item We recognize that reproducibility may be tricky in some cases, in which case authors are welcome to describe the particular way they provide for reproducibility. In the case of closed-source models, it may be that access to the model is limited in some way (e.g., to registered users), but it should be possible for other researchers to have some path to reproducing or verifying the results.
        \end{enumerate}
    \end{itemize}

\item {\bf Open access to data and code}
    \item[] Question: Does the paper provide open access to the data and code, with sufficient instructions to faithfully reproduce the main experimental results, as described in supplemental material?
    \item[] Answer: \answerNo{}
    \item[] Justification: The datasets, baseline methods, and models used in the paper are fully open-source and available on Hugging Face. The paper includes the key implementation steps and code in Section 4 and the Appendix D.
    \item[] Guidelines:
    \begin{itemize}
        \item The answer \answerNA{} means that paper does not include experiments requiring code.
        \item Please see the NeurIPS code and data submission guidelines (\url{https://neurips.cc/public/guides/CodeSubmissionPolicy}) for more details.
        \item While we encourage the release of code and data, we understand that this might not be possible, so \answerNo{} is an acceptable answer. Papers cannot be rejected simply for not including code, unless this is central to the contribution (e.g., for a new open-source benchmark).
        \item The instructions should contain the exact command and environment needed to run to reproduce the results. See the NeurIPS code and data submission guidelines (\url{https://neurips.cc/public/guides/CodeSubmissionPolicy}) for more details.
        \item The authors should provide instructions on data access and preparation, including how to access the raw data, preprocessed data, intermediate data, and generated data, etc.
        \item The authors should provide scripts to reproduce all experimental results for the new proposed method and baselines. If only a subset of experiments are reproducible, they should state which ones are omitted from the script and why.
        \item At submission time, to preserve anonymity, the authors should release anonymized versions (if applicable).
        \item Providing as much information as possible in supplemental material (appended to the paper) is recommended, but including URLs to data and code is permitted.
    \end{itemize}

\item {\bf Experimental setting/details}
    \item[] Question: Does the paper specify all the training and test details (e.g., data splits, hyperparameters, how they were chosen, type of optimizer) necessary to understand the results?
    \item[] Answer: \answerYes{} 
    \item[] Justification: The paper provides a detailed description of the experimental data setup and hyperparameter settings in Section 5 and the Appendix A.
    \item[] Guidelines:
    \begin{itemize}
        \item The answer \answerNA{} means that the paper does not include experiments.
        \item The experimental setting should be presented in the core of the paper to a level of detail that is necessary to appreciate the results and make sense of them.
        \item The full details can be provided either with the code, in appendix, or as supplemental material.
    \end{itemize}

\item {\bf Experiment statistical significance}
    \item[] Question: Does the paper report error bars suitably and correctly defined or other appropriate information about the statistical significance of the experiments?
    \item[] Answer: \answerYes{}
    \item[] Justification:  All results are averaged over multiple tests, and we report the mean accuracy along with the standard deviation.
    \item[] Guidelines:
    \begin{itemize}
        \item The answer \answerNA{} means that the paper does not include experiments.
        \item The authors should answer \answerYes{} if the results are accompanied by error bars, confidence intervals, or statistical significance tests, at least for the experiments that support the main claims of the paper.
        \item The factors of variability that the error bars are capturing should be clearly stated (for example, train/test split, initialization, random drawing of some parameter, or overall run with given experimental conditions).
        \item The method for calculating the error bars should be explained (closed form formula, call to a library function, bootstrap, etc.)
        \item The assumptions made should be given (e.g., Normally distributed errors).
        \item It should be clear whether the error bar is the standard deviation or the standard error of the mean.
        \item It is OK to report 1-sigma error bars, but one should state it. The authors should preferably report a 2-sigma error bar than state that they have a 96\% CI, if the hypothesis of Normality of errors is not verified.
        \item For asymmetric distributions, the authors should be careful not to show in tables or figures symmetric error bars that would yield results that are out of range (e.g., negative error rates).
        \item If error bars are reported in tables or plots, the authors should explain in the text how they were calculated and reference the corresponding figures or tables in the text.
    \end{itemize}

\item {\bf Experiments compute resources}
    \item[] Question: For each experiment, does the paper provide sufficient information on the computer resources (type of compute workers, memory, time of execution) needed to reproduce the experiments?
    \item[] Answer: \answerYes{} 
    \item[] Justification: We report them in the Appendix A.
    \item[] Guidelines:
    \begin{itemize}
        \item The answer \answerNA{} means that the paper does not include experiments.
        \item The paper should indicate the type of compute workers CPU or GPU, internal cluster, or cloud provider, including relevant memory and storage.
        \item The paper should provide the amount of compute required for each of the individual experimental runs as well as estimate the total compute. 
        \item The paper should disclose whether the full research project required more compute than the experiments reported in the paper (e.g., preliminary or failed experiments that didn't make it into the paper). 
    \end{itemize}
    
\item {\bf Code of ethics}
    \item[] Question: Does the research conducted in the paper conform, in every respect, with the NeurIPS Code of Ethics \url{https://neurips.cc/public/EthicsGuidelines}?
    \item[] Answer: \answerYes{} 
    \item[] Justification: The discussion of the ethics and impact can be consulted in the Appendix F.
    \item[] Guidelines:
    \begin{itemize}
        \item The answer \answerNA{} means that the authors have not reviewed the NeurIPS Code of Ethics.
        \item If the authors answer \answerNo, they should explain the special circumstances that require a deviation from the Code of Ethics.
        \item The authors should make sure to preserve anonymity (e.g., if there is a special consideration due to laws or regulations in their jurisdiction).
    \end{itemize}

\item {\bf Broader impacts}
    \item[] Question: Does the paper discuss both potential positive societal impacts and negative societal impacts of the work performed?
    \item[] Answer: \answerYes{} 
    \item[] Justification: The discussion of the broader impacts can be consulted in the Appendix F.
    \item[] Guidelines:
    \begin{itemize}
        \item The answer \answerNA{} means that there is no societal impact of the work performed.
        \item If the authors answer \answerNA{} or \answerNo, they should explain why their work has no societal impact or why the paper does not address societal impact.
        \item Examples of negative societal impacts include potential malicious or unintended uses (e.g., disinformation, generating fake profiles, surveillance), fairness considerations (e.g., deployment of technologies that could make decisions that unfairly impact specific groups), privacy considerations, and security considerations.
        \item The conference expects that many papers will be foundational research and not tied to particular applications, let alone deployments. However, if there is a direct path to any negative applications, the authors should point it out. For example, it is legitimate to point out that an improvement in the quality of generative models could be used to generate Deepfakes for disinformation. On the other hand, it is not needed to point out that a generic algorithm for optimizing neural networks could enable people to train models that generate Deepfakes faster.
        \item The authors should consider possible harms that could arise when the technology is being used as intended and functioning correctly, harms that could arise when the technology is being used as intended but gives incorrect results, and harms following from (intentional or unintentional) misuse of the technology.
        \item If there are negative societal impacts, the authors could also discuss possible mitigation strategies (e.g., gated release of models, providing defenses in addition to attacks, mechanisms for monitoring misuse, mechanisms to monitor how a system learns from feedback over time, improving the efficiency and accessibility of ML).
    \end{itemize}
    
\item {\bf Safeguards}
    \item[] Question: Does the paper describe safeguards that have been put in place for responsible release of data or models that have a high risk for misuse (e.g., pre-trained language models, image generators, or scraped datasets)?
    \item[] Answer: \answerNA{} 
    \item[] Justification: This paper presents an improved approach based on the existing model architecture. The paper poses no such risks.
    \item[] Guidelines:
    \begin{itemize}
        \item The answer \answerNA{} means that the paper poses no such risks.
        \item Released models that have a high risk for misuse or dual-use should be released with necessary safeguards to allow for controlled use of the model, for example by requiring that users adhere to usage guidelines or restrictions to access the model or implementing safety filters. 
        \item Datasets that have been scraped from the Internet could pose safety risks. The authors should describe how they avoided releasing unsafe images.
        \item We recognize that providing effective safeguards is challenging, and many papers do not require this, but we encourage authors to take this into account and make a best faith effort.
    \end{itemize}

\item {\bf Licenses for existing assets}
    \item[] Question: Are the creators or original owners of assets (e.g., code, data, models), used in the paper, properly credited and are the license and terms of use explicitly mentioned and properly respected?
    \item[] Answer: \answerYes{} 
    \item[] Justification:  This article uses assets reasonably in compliance with the license, and the assets used are cited in the article.
    \item[] Guidelines:
    \begin{itemize}
        \item The answer \answerNA{} means that the paper does not use existing assets.
        \item The authors should cite the original paper that produced the code package or dataset.
        \item The authors should state which version of the asset is used and, if possible, include a URL.
        \item The name of the license (e.g., CC-BY 4.0) should be included for each asset.
        \item For scraped data from a particular source (e.g., website), the copyright and terms of service of that source should be provided.
        \item If assets are released, the license, copyright information, and terms of use in the package should be provided. For popular datasets, \url{paperswithcode.com/datasets} has curated licenses for some datasets. Their licensing guide can help determine the license of a dataset.
        \item For existing datasets that are re-packaged, both the original license and the license of the derived asset (if it has changed) should be provided.
        \item If this information is not available online, the authors are encouraged to reach out to the asset's creators.
    \end{itemize}

\item {\bf New assets}
    \item[] Question: Are new assets introduced in the paper well documented and is the documentation provided alongside the assets?
    \item[] Answer: \answerNo{} 
    \item[] Justification: The paper does not release new assets.
    \item[] Guidelines:
    \begin{itemize}
        \item The answer \answerNA{} means that the paper does not release new assets.
        \item Researchers should communicate the details of the dataset\slash code\slash model as part of their submissions via structured templates. This includes details about training, license, limitations, etc. 
        \item The paper should discuss whether and how consent was obtained from people whose asset is used.
        \item At submission time, remember to anonymize your assets (if applicable). You can either create an anonymized URL or include an anonymized zip file.
    \end{itemize}

\item {\bf Crowdsourcing and research with human subjects}
    \item[] Question: For crowdsourcing experiments and research with human subjects, does the paper include the full text of instructions given to participants and screenshots, if applicable, as well as details about compensation (if any)? 
    \item[] Answer: \answerNA{} 
    \item[] Justification: The paper does not involve crowdsourcing nor research with human subjects.
    \item[] Guidelines:
    \begin{itemize}
        \item The answer \answerNA{} means that the paper does not involve crowdsourcing nor research with human subjects.
        \item Including this information in the supplemental material is fine, but if the main contribution of the paper involves human subjects, then as much detail as possible should be included in the main paper. 
        \item According to the NeurIPS Code of Ethics, workers involved in data collection, curation, or other labor should be paid at least the minimum wage in the country of the data collector. 
    \end{itemize}

\item {\bf Institutional review board (IRB) approvals or equivalent for research with human subjects}
    \item[] Question: Does the paper describe potential risks incurred by study participants, whether such risks were disclosed to the subjects, and whether Institutional Review Board (IRB) approvals (or an equivalent approval/review based on the requirements of your country or institution) were obtained?
    \item[] Answer: \answerNA{} 
    \item[] Justification: The paper does not involve crowdsourcing nor research with human subjects.
    \item[] Guidelines:
    \begin{itemize}
        \item The answer \answerNA{} means that the paper does not involve crowdsourcing nor research with human subjects.
        \item Depending on the country in which research is conducted, IRB approval (or equivalent) may be required for any human subjects research. If you obtained IRB approval, you should clearly state this in the paper. 
        \item We recognize that the procedures for this may vary significantly between institutions and locations, and we expect authors to adhere to the NeurIPS Code of Ethics and the guidelines for their institution. 
        \item For initial submissions, do not include any information that would break anonymity (if applicable), such as the institution conducting the review.
    \end{itemize}

\item {\bf Declaration of LLM usage}
    \item[] Question: Does the paper describe the usage of LLMs if it is an important, original, or non-standard component of the core methods in this research? Note that if the LLM is used only for writing, editing, or formatting purposes and does \emph{not} impact the core methodology, scientific rigor, or originality of the research, declaration is not required.
    \item[] Answer: \answerNA{} 
    \item[] Justification: The core method development in this paper does not involve LLMs as any important, original, or non-standard components.
    \item[] Guidelines:
    \begin{itemize}
        \item The answer \answerNA{} means that the core method development in this research does not involve LLMs as any important, original, or non-standard components.
        \item Please refer to our LLM policy in the NeurIPS handbook for what should or should not be described.
    \end{itemize}

\end{enumerate}

\end{document}